\documentclass[lettersize,journal]{IEEEtran}
\usepackage{amsmath,amsfonts}
\usepackage{array}
\usepackage[caption=false,font=normalsize,labelfont=sf,textfont=sf]{subfig}
\usepackage{textcomp}
\usepackage{stfloats}
\usepackage{url}
\usepackage{verbatim}
\usepackage{graphicx}
\usepackage{cite}
\usepackage{multirow}
\usepackage[normalem]{ulem}
\useunder{\uline}{\ul}{}
\usepackage[table,xcdraw]{xcolor}
\usepackage{pifont}
\usepackage{ragged2e}
\newcommand{\pub}[1]{{\color{gray}{\footnotesize{[{#1}]}}}}
\newcommand{\cmark}{\color{black}{\ding{51}}}
\definecolor{c1}{RGB}{255, 0, 0}
\definecolor{c2}{RGB}{0, 255, 0}
\definecolor{c3}{RGB}{0, 0, 255}
\definecolor{c4}{RGB}{255, 255, 0}
\definecolor{c5}{RGB}{255, 0, 255}
\definecolor{c6}{RGB}{0, 255, 255}
\definecolor{c7}{RGB}{200, 100, 0}
\definecolor{c8}{RGB}{0, 200, 100}
\definecolor{c9}{RGB}{100, 0, 200}
\definecolor{c10}{RGB}{200, 0, 100}
\definecolor{c11}{RGB}{100, 200, 0}
\definecolor{c12}{RGB}{0, 100, 200}
\definecolor{c13}{RGB}{150, 75, 75}
\definecolor{c14}{RGB}{75, 150, 75}
\definecolor{c15}{RGB}{75, 75, 150}
\definecolor{c16}{RGB}{255, 100, 100}
\definecolor{c17}{RGB}{100, 255, 100}
\definecolor{c18}{RGB}{100, 100, 255}
\definecolor{c19}{RGB}{255, 150, 75}
\definecolor{c20}{RGB}{75, 255, 150}
\definecolor{c21}{RGB}{150, 75, 255}
\definecolor{c22}{RGB}{50, 50, 50}

\usepackage[ruled]{algorithm2e}
\usepackage[capitalize]{cleveref}
\crefname{section}{Sec.}{Secs.}
\Crefname{section}{Section}{Sections}
\Crefname{table}{Table}{Tables}
\crefname{table}{Tab.}{Tabs.}
\usepackage{caption}
\newcommand{\blue}{\textcolor{black}}

\begin{document}

\title{MambaHSI: Spatial-Spectral Mamba for Hyperspectral Image Classification}

\author{Yapeng Li,~\IEEEmembership{Graduate Student Member,~IEEE,} Yong Luo,~\IEEEmembership{Member,~IEEE,} Lefei Zhang,~\IEEEmembership{Senior Member,~IEEE,} Zengmao Wang,~\IEEEmembership{Member,~IEEE,} Bo Du,~\IEEEmembership{Senior Member,~IEEE,}

\thanks{This work was supported by the National Natural Science Foundation of China under Grants 62225113 and U23A20318, the Innovative Research Group Project of Hubei Province under Grants 2024AFA017.
\textit{(Corresponding authors: Zengmao Wang, Bo Du.)}}

\thanks{Y. Li, Y. Luo, L. Zhang, Z. Wang, and B. Du are with the National Engineering Research Center for Multimedia Software, School of Computer Science, Institute of Artificial Intelligence, and Hubei Key Laboratory of Multimedia and Network Communication Engineering, Wuhan University, Wuhan 430072, China (e-mail: yapengli@whu.edu.cn; luoyong@whu.edu.cn; zhanglefei@whu.edu.cn; wangzengmao@whu.edu.cn; dubo@whu.edu.cn).}
}

\markboth{IEEE TRANSACTIONS ON GEOSCIENCE AND REMOTE SENSING}
{Shell \MakeLowercase{\textit{et al.}}: }


\maketitle

\begin{abstract}
Transformer has been extensively explored for hyperspectral image classification. However,
transformer poses challenges in terms of speed and memory usage because of its quadratic computational complexity. 
Recently the Mamba model
has emerged as a promising approach, which has strong long-distance modeling capabilities while maintaining a linear computational complexity. However, representing the hyperspectral image is challenging for the Mamba due to the requirement for an integrated spatial and spectral understanding.
To remedy these drawbacks, we propose a novel hyperspectral image classification model based on a Mamba model, named MambaHSI, which can simultaneously model long-range interaction of the whole image and integrate spatial and spectral information in an adaptive manner. Specifically, we design a spatial Mamba block to model the long-range interaction of the whole image at the pixel-level. Then we propose a spectral Mamba block to split the spectral vector into multiple groups, mine the relations across different spectral groups, and extract spectral features. Finally, we propose a spatial-spectral fusion module to adaptively integrate spatial and spectral features of a hyperspectral image. To our best knowledge, this is the first image-level hyperspectral image classification model based on the Mamba. We conduct extensive experiments on four diverse hyperspectral image datasets. The results demonstrate the effectiveness and superiority of the proposed model for hyperspectral image classification. This reveals the great potential of Mamba to be the next-generation backbone for hyperspectral image models. Codes are available at \url{https://github.com/li-yapeng/MambaHSI}.

\end{abstract}

\section{Introduction}
\IEEEPARstart{W}ITH the development of earth observation technology, hyperspectral image (HSI) can be widely acquired~\cite{WideUseHSI}. Different from the traditional vision system that capture images only with the RGB channels, hyperspectral images can cover a much larger spectral range from the visible spectrum, near-infrared, mid-infrared to far-infrared with dozens or even hundreds of continuous wavebands~\cite{Bands_HSI_plaza,SSFCN,YuxiangTIP16}. Benefiting from the abundant and detailed spectral information, HSI provides an opportunity to identify objects which are difficult to distinguish in natural RGB images~\cite{HSI_VS_RGB_Benefit}. This makes more accurate observations of the Earth possible~\cite{SSFCN}. Owing to the advantages mentioned above, HSIs are widely applied to diverse applications~\cite{HSI_DLreview}, such as urban mapping, resource exploring, environmental monitoring, and so on. As the fundamental task of these applications, HSI classification is to predict the class label of each pixel in the image~\cite{DefHSIClassification}.

\begin{figure}
    \centering
    \includegraphics[width=\linewidth]{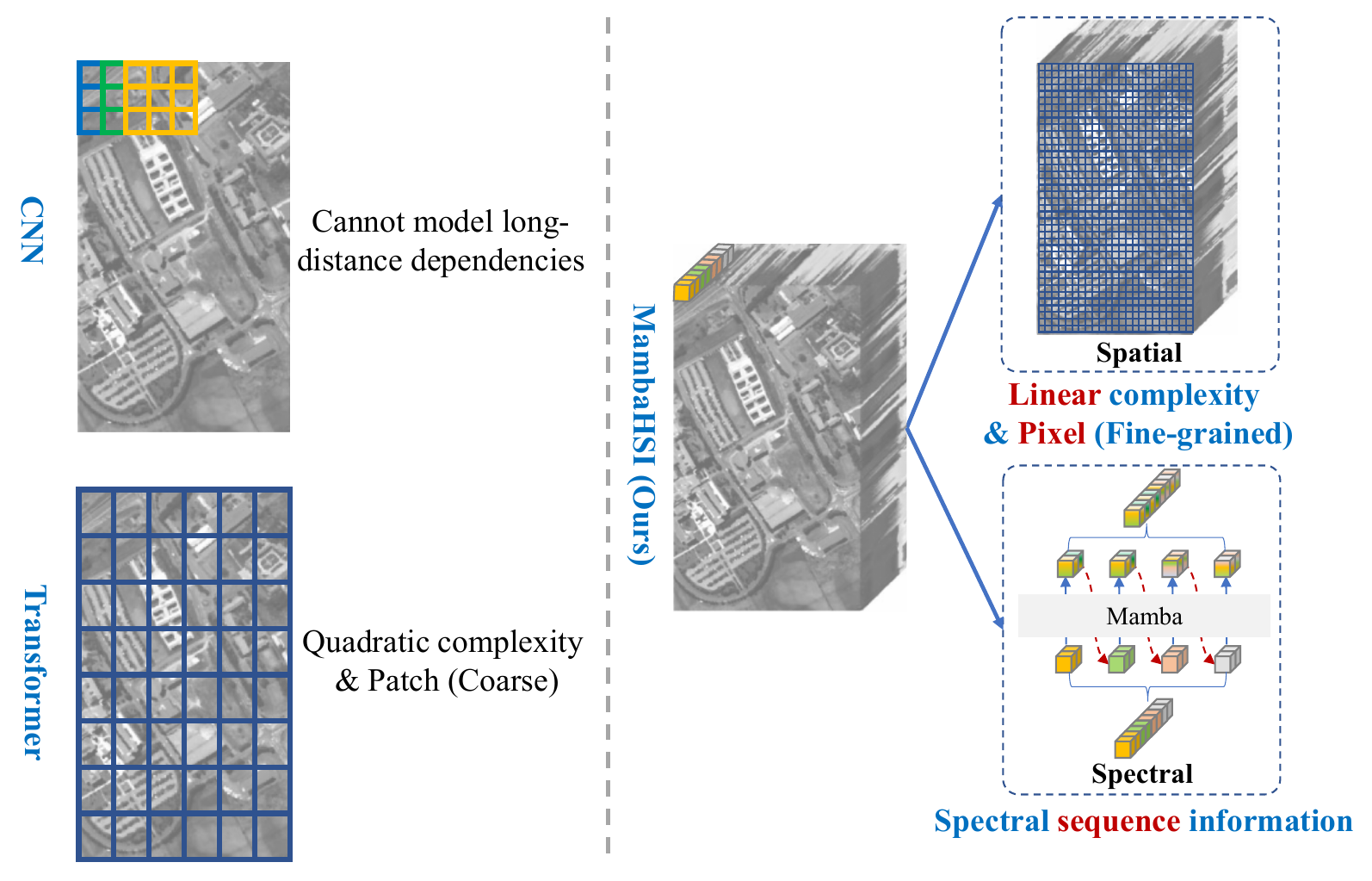}
    \caption{\small{Motivation illustration. The local characteristics of CNNs and the quadratic complexity of Transformers limit their ability to achieve fine-grained global modeling. In contrast, the proposed MambaHSI model can achieve pixel-level fine-grained spatial feature modeling with linear complexity. By incorporating spectral sequence information, MambaHSI enhances the extraction of spectral features.}}
    \vspace{-20pt}
    \label{fig:enter-label}
\end{figure}

Benefiting from the development of deep learning technology, hyperspectral image classification has witnessed a rapid advancement in the past few years~\cite{HSI_DLreview,fu2024we}.
\blue{Hyperspectral image classification methods can be generally divided into traditional machine learning (ML-based)~\cite{SVM_Comp,RF_Comp} and deep learning (DL-based)~\cite{hong2020graph,GSC-Vit_TGRS24}. DL-based can be roughly fall into Graph Convolutional Network (GCN-based)~\cite{hong2020graph,DMSGer}, Convolutional Neural Network (CNN-based)~\cite{fu2024resc,CLOLN}, and Transformer-based~\cite{spectralformer,GSC-Vit_TGRS24} methods.}
However, the mainstream HSI classification methods based on CNN and Transformer have inherent limitations. CNN-based models are constrained by their local receptive field, which hinders their ability to model long-range dependencies~\cite{vm-unet}. Although Transformer-based models show superior performance for global modeling, the self-attention mechanism of the transformer layer demands quadratic complexity in terms of the sequence length~\cite{transformer}.
\blue{This results in a high computational burden~\cite{VisionMamba} and impedes the extraction of fine-grained spatial features at the pixel level.}
The drawbacks of these models compel us to develop a novel framework for HSI classification, capable of modeling strong long-range dependencies and maintaining linear computational complexity.
 
Recently, state space models (SSMs)~\cite{gu_PHD_Paper, gu2021combining}, in particular structured state space sequence models (S4)~\cite{gu2022S4}, have emerged as an efficient and effective layer to construct deep networks, and have obtained superior performance in continuous long-sequence data analysis~\cite{goel2022s,gu2022S4}. Mamba~\cite{gu2023mamba} further improved S4 with a selective mechanism, which enables the model to select relevant information in an input-dependent manner. Besides, Mamba achieves higher computational efficiency than Transformer by combining with hardware-aware implementation~\cite{gu2023mamba}.
Benefiting from the strong long-range modeling capabilities while maintaining linear computational complexity, the Mamba has received substantial research across many fields, such as language understanding~\cite{gu2022S4,gu2023mamba}, medical image analysis~\cite{U-mamba,vm-unet}, computer vision~\cite{VisionMamba,mambasurvey_xu}, and so on~\cite{oshima2024ssm,rsmamba2024,pointmamba}. \blue{However, the aforementioned methods cannot be applied to hyperspectral images because these methods ignore the spectral continuity of hyperspectral images.}

In this paper, we propose a novel hyperspectral image classification framework based on Mamba, termed as MambaHSI, which not only models the long-range dependencies but also comprehensively utilizes the spatial and spectral information. Specifically, MambaHSI is a pure-SSM-based model, which takes the whole image as inputs and can be trained in an end-to-end manner. The pure-SSM-based structure enables model to capture the long-range dependencies while maintaining the linear computational complexity. The end-to-end training method enables the model to be optimized as a whole to obtain better parameters. Then, considering the spatial and spectral characteristics of hyperspectral images, we design a spatial and spectral Mamba block to extract spatial and spectral information respectively. Benefiting from the strong long-range modeling ability, the spatial and spectral Mamba block can capture more discriminative spatial and spectral features for classification. To comprehensively exploit the spatial and spectral information, we propose a spatial-spectral fusion module, which can adaptively fuse the spatial and spectral information based on the importance of spatial and spectral. The idea of residual learning has also been introduced to help with model learning.

To Summarize, the main contributions of this paper are:
\begin{itemize}
    \item To our best knowledge, the proposed MambaHSI is the first image-level hyperspectral image classification model based on SSM, which can simultaneously model long-range interaction of whole image and integrate spatial and spectral image information.
    \item We design a spatial and spectral mamba block to extract the spatial and spectral information respectively. Benefiting from the strong long-range modeling capability of mamba, the proposed spatial and spectral mamba block can model long-range interaction of whole image.
    \item We propose a spatial-spectral fusion module, which can adaptively estimate the importance of spatial and spectral information to guide their fusion. Besides, the residual learning idea is introduced to help with the module training. 
\end{itemize}

We conduct extensive experiments on four diverse real-world hyperspectral image datasets. The results show that our method achieves superior performance, surpassing the state-of-the-art CNN-based and Transformer-based hyperspectral image classification models.
\section{Related Work}
\subsection{Hyperspectral Image Classification}
\blue{Hyperspectral image classification methods typically fall into ML-based~\cite{tu2018knn,RF_Comp} and DL-based~\cite{DMSGer,CLOLN}. DL-based be roughly divided into GCN-based~\cite{hong2020graph,DMSGer}, CNN-based~\cite{CVSSN,CLOLN}, and Transformer-based~\cite{SSFTT_TGRS22,GAHT} methods.}

\blue{\textbf{ML-based.} 
The early approaches adopt traditional machine learning, such as random forest~\cite{RF_Comp}, and support vector machine (SVM)~\cite{SVM_Comp}, to classify the spectral features. However, the performance of these methods is limited because of lacking surrounding information. That is, the above methods utilize only the spectral information of a single pixel, neglecting the spatial information from surrounding neighboring pixels, which leads to suboptimal performance~\cite{SSFCN}. To remedy this problem, spatial information is incorporated into consideration. For example, the superpixel segmentation~\cite{SuperPixel2015}, the extended morphological profiles~\cite{morphological2005}, and multiple kernel learning~\cite{mkl2012} are adopted to generate discriminative spectral-spatial features for classification. 
However, these methods mainly classify image using shallow features, so that the model cannot characterize the essential attribute of the objects~\cite{represent_learning_review}. This hinders the performance of classification.}

\blue{\textbf{GCN-based.} GCNs can be used to model long-range spatial relations in the HS image~\cite{hong2020graph}. Qin et al.~\cite{qin2018spectral} simultaneously consider spatial and spectral neighborhoods by extending original GCN to second-order GCN. Wan et al.~\cite{Wan_2021_GCN} utilized superpixel segmentation on HS image and fed it into GCN. The GiGCN model~\cite{GiGCN} from a superpixel viewpoint can fully exploit information inside and outside superpixels. Dynamic multi-scale graph convolutional network classifier (DMSGer) simultaneously captures pixel-level and region-level features to boost classification performance~\cite{DMSGer}. Due to the GCNs’ limitations in the graph construction, particularly for large graphs (need expensive computational cost), GCNs fail to classify or identify materials in large-scale hyperspectral scenes~\cite{GCN-ref-review}.
}

\textbf{CNN-based.} CNN can automatically capture features from shallow to deep level by stacking layer by layer, thus attracting much attention in the field of hyperspectral image classification~\cite{yang2023center}. ~\cite{CNN1D_Spectral} extracts spectral features via a five-layer 1D CNN. 
Considering the differences in scale of ground objects, ~\cite{MSCNN_patch} generates multi-scale image patches for each pixel, and then extracts the multi-scale features of the generated multi-scale patches for classification. Furthermore, ~\cite{Spa-Spe_3DCNN} jointly captures the spatial and spectral features via 3D-CNN. Different from the above patch-level methods, many image-level methods have been proposed to reduce running time. SSFCN~\cite{SSFCN} is the first image-level HSI classification method, which is a double-branch FCN network to separately extract spectral and spatial features. Next, multi-scale strategy and attention mechanism are introduced into FCN, which improve the multi-scale feature extraction and fusion capabilities~\cite{wang2021fully}. 
However, these CNN-based methods cannot model long-range dependencies due to the inherent locality of CNN, which hinders the feature extraction ability of models~\cite{VisionMamba}.

\textbf{Transformer-based.} Recently, the Transformer have been widely used in HSI classification because of their abilities to model long-range dependencies and extract global spatial features. SpectralFormer~\cite{spectralformer} learns sequence information from local signatures with a transformer. Nevertheless, SpectralFormer ignores the spatial location information. To this end, 
some spatial-spectral transformers~\cite{spatial-spectral_former1_TGRS22,SSFTT_TGRS22} have been proposed to simultaneously capture spectral and spatial information. Unfortunately, these methods do not consider the differences in scales of different ground objects, limiting the ability to identify multi-scale objects. In ~\cite{CSIL_ISPRS23}, a center-to-surrounding interactive learning (CSIL) framework based on the Transformer has been proposed to extract multi-scale features for classification. ~\cite{GSC-Vit_TGRS24} pointed out that the transformer ignores the effective representation of local features due to the inherent modeling global correlation, and then the GSC-ViT proposed in ~\cite{GSC-Vit_TGRS24} can capture local spectral–spatial information in hyperspectral image.
\blue{Xue et al.~\cite{xue2022local} proposed a local transformer with spatial partition restore network (SPRLT-Net) to model locally detailed spatial discrepancies. Li et al.~\cite{SPFormer} proposed a lightweight Transformer architecture model, SPFormer, for few-shot HSI classification, which effectively reduces the parameters of transformer.} However, the complexity of the self-attention mechanism is quadratic~\cite{transformer}. This poses challenges in terms of speed and memory usage, and hinders the model’s ability to model long-range dependencies~\cite{VisionMamba}.

In general, existing hyperspectral image classification methods have limited ability to model the long-range dependencies because of its inherent properties, such as the locality of CNN and the quadratic complexity. Different from existing HSI classification methods, MambaHSI is the first Mamba-based HSI classification method that takes whole HSI image as the model's input, which can model long-range dependencies while maintaining linear computational complexity.

\subsection{State Space Models}
Recent research advancements have sparked a surge of interest in the state space model (SSM). SSM originates from the classic Kalman Filter model, and modern SSMs are good at capturing long-range dependencies and can efficiently parallelize calculations~\cite{VisionMamba}. SSMs are a novel alternative to CNNs or Transformers. 

\textbf{State space models for long sequence.} The Structured State-Space Sequence (S4)~\cite{gu2022S4} model can model long-range dependencies. The promising property of linear complexity for sequence length attracts further exploration. ~\cite{S5} designs a S5 layer by introducing MIMO SSM and efficient parallel scan into S4 layer. ~\cite{H3} fills the performance gap between SSMs and Transformers in language modeling. ~\cite{GateS4} introduces more gating units to improve the expressivity of S4. \blue{Recently, Mamba outperforms Transformers at various large-scale real data and maintains linear scaling in sequence length~\cite{gu2023mamba}.} 

\textbf{State space models for visual applications.} ~\cite{S4ND} extends 1D S4 to handle 2D images and 3D videos. TranS4mer~\cite{TranS4mer} combines the strengths of S4 and self-attention, and achieves state-of-the-art performance for movie scene detection. ~\cite{wang2023selective} introduces a selectivity mechanism to S4, significantly improving the performance of S4 on long-form video understanding with lower memory footprint. U-Mamba~\cite{U-mamba} combines Mamba with U-shaped architecture for biomedical image segmentation. 

\blue{\textbf{State space models for HSI classification.} Recently,~\cite{yao2024spectralmamba,he20243dss,huang2024spectral,sheng2024dualmamba} adopt Mamba for hyper-spectral image classification. SpectralMamba~\cite{yao2024spectralmamba} introduced PSS and GSSM module to ease the sequentially learning in the state domain and rectify the spectrum, respectively. Huang et al.~\cite{huang2024spectral} proposed a spectral-spatial Mamba (3DSS-Mamba) for HSI classification. He et al.~\cite{he20243dss} explored the application of Mamba to hyperspectral blocks in 3D from a 3D perspective. However, the aforementioned methods utilize hyperspectral patches as input. This results in a significant amount of redundant computation when inferring the entire image, impeding practical deployment and application. Additionally, the fixed size of the patch inputs restricts the model's ability to fully exploit the image information, thereby limiting its feature extraction capacity.
}

To this end, we first design a spatial and spectral Mamba block to capture spatial and spectral information, respectively. Besides, we propose a spatial-spectral fusion module to adaptively fuse the spatial and spectral information. This enables the proposed MambaHSI to fully integrate both spatial and spectral information while modeling long-range dependencies in a linear complexity manner.

\begin{figure*}[!t]
    \centering 
    \includegraphics[width=\linewidth]{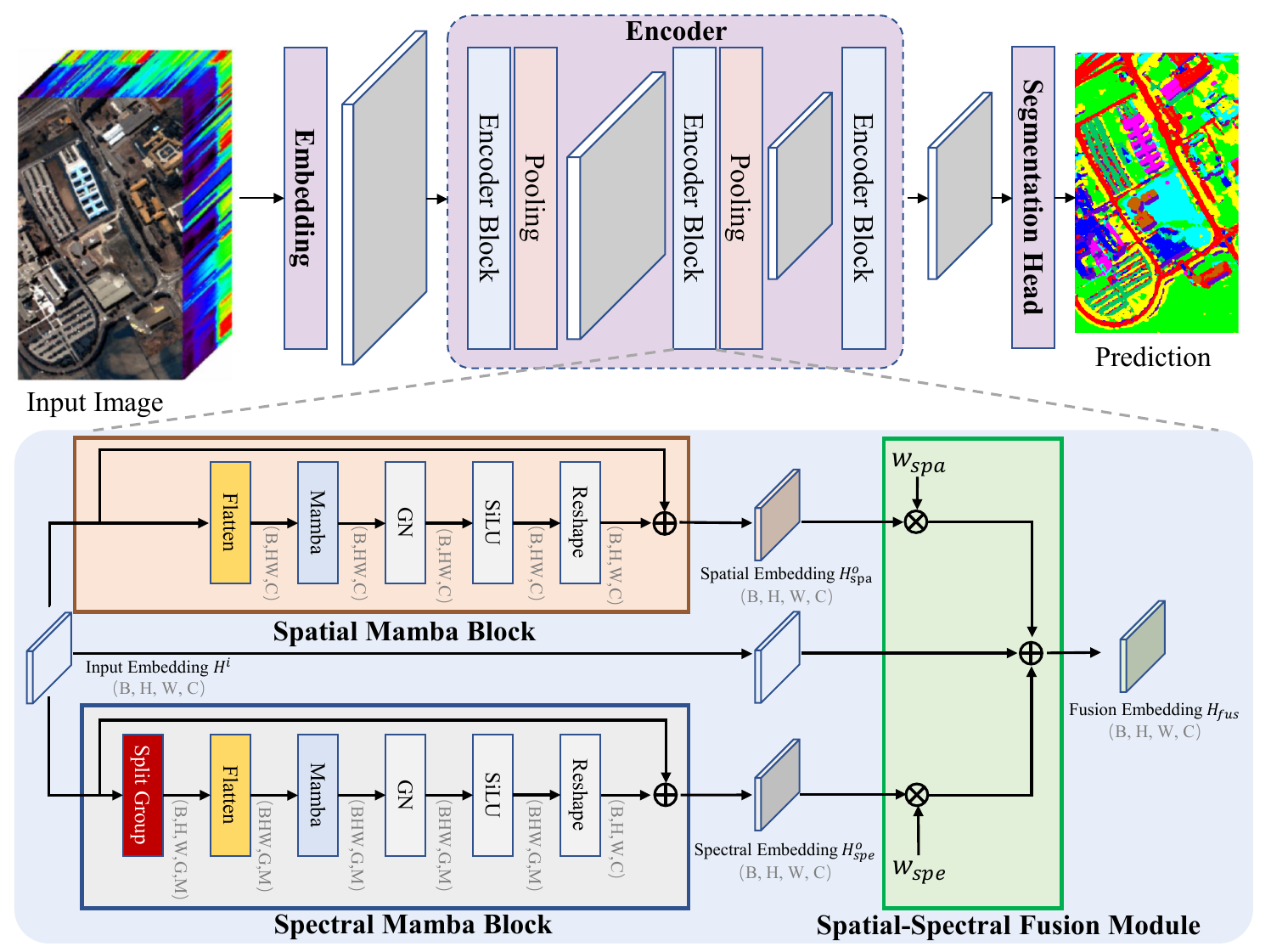}
    \vspace{-15pt}
    \caption{\small{Overview of the proposed MambaHSI framework. The whole hyperspectral image is fed into the  embedding layer to obtain pixel-level embeddings. Then these embeddings are taken as the inputs of the encoder to model the long-range dependencies and capture the discriminative features. Finally, the segmentation head classifies the features extracted by the encoder to obtain the final prediction. The encoder block contains three components: spatial Mamba block for capturing spatial features, spectral Mamba block for extracting spectral features, and spatial-spectral fusion module to fuse the spatial and spectral features.}}
    \vspace{-15pt}
    \label{fig:MambaHSI_framework}
\end{figure*}

\section{Preliminaries}
The modern SSM-based models, \textit{i.e.}, Structure State Space Sequence Models (S4) and Mamba, both are inspired by a classical continuous system that maps a 1-D function or sequence, denoted as $x(t) \in \mathcal{R}$, to an output $y(t) \in \mathcal{R}$ through a hidden state $h(t) \in \mathcal{R}^{N}$. The process can be represented as a linear Ordinary Differential Equation (ODE):
\begin{equation}
\label{eq:continuousSSM}
    \begin{split}
        &h'(t)=\mathbf{A}h(t) + \mathbf{B}x(t), \\
        &y(t) = \mathbf{C}h(t),
    \end{split}
\end{equation}
where $\mathbf{A} \in \mathcal{R}^{N \times N}$ denotes the state matrix, $\mathbf{B} \in \mathcal{R}^{N \times 1}$ and $\mathbf{C} \in \mathcal{R}^{N \times 1}$ signify the projection parameters.

The S4 and Mamba discretize the continuous system to make it more suitable for deep learning scenarios. Specifically, they introduce a timescale parameter $\mathbf{\Delta}$ and transform the continuous parameters $\mathbf{A}$ and $\mathbf{B}$ into discrete parameters $\overline{\mathbf{A}}$ and $\overline{\mathbf{B}}$ using a fixed discretization rule. The commonly used discretization rule is zero-order hold (ZOH), which can be defined as follows:
\begin{equation}
    \begin{split}
        & \overline{\mathbf{A}} = \exp(\mathbf{\Delta \mathbf{A}}), \\
        & \overline{\mathbf{B}} = (\mathbf{\Delta}\mathbf{A})^{-1}(\exp(\mathbf{\Delta\mathbf{A}})-\mathbf{I})\cdot \mathbf{\Delta}\mathbf{B}.
    \end{split}
\end{equation}

After the discretization of $\overline{\mathbf{A}},\overline{\mathbf{B}}$, the SSM-based models can be computered in two ways: linear recurrence or global convolution, defined as ~\cref{eq:linear_recurrence} and ~\cref{eq:global_conv}, respectively.
\begin{equation}
\label{eq:linear_recurrence}
    \begin{split}
         & h_{t} = \overline{\mathbf{A}}h_{t-1} + \overline{\mathbf{B}}x_{t}, \\
         & y_{t} = \mathbf{C}h_{t}.
    \end{split}
\end{equation}

\begin{equation}
    \label{eq:global_conv}
    \begin{split}
        & \overline{\mathbf{K}}=(\mathbf{C}\overline{\mathbf{B}},\mathbf{C}\overline{\mathbf{A}\mathbf{B}},\dots,\mathbf{C}\overline{\mathbf{A}}^{L-1}\overline{\mathbf{B}}),  \\
        & y = x * \overline{\mathbf{K}},
    \end{split}
\end{equation}
where $L$ and $\overline{\mathbf{K}} \in \mathcal{R}^{L}$ denote the length of the input sequence $x$ and a structured convolutional kernel, respectively.

\section{Methodology}
\subsection{Motivation and Overall Idea}
\blue{Hyperspectral image classification is a pixel-level classification task~\cite{DefHSIClassification}. This means that HSI classification models need to learn features that can finely distinguish subtle differences between pixels. Inspired by \cite{nonlocal}, the long-range modeling capability is vital for capturing discriminative features. Similar observations have been made in the HSI classification field \cite{NonLocal_HSI,LongRange_transformer23,LongRange_23}. Unfortunately, existing HSI classification methods have limited ability to model long-range dependencies due to inherent limitations, such as the locality of CNNs and the quadratic computational complexity of Transformers. Because of the quadratic computational complexity of Transformers, models based on Transformers often use patches as the basic token units when processing images, which hinders the model's ability to extract finer pixel-level representations for classification. This motivates us to develop a new framework that can model long-range dependencies with linear complexity, enabling finer feature extraction at the pixel level for classification. Therefore, we introduce Mamba as the basic unit to model long-range dependencies, which is vital for obtaining discriminative pixel-wise spatial features. However, Mamba ignores the fact that hyperspectral (HS) images are characterized by approximately contiguous spectral information. Therefore, we rethink HS image classification from a sequential perspective with Mamba and propose a spectral Mamba block to utilize the sequential nature of the spectrum. To comprehensively capture the spatial-spectral information, we propose a spatial-spectral fusion module to adaptively fuse the spatial and spectral information, improving the recognizability of the extracted features.}

\subsection{Overview}
An overview of the proposed MambaHSI is shown in ~\cref{fig:MambaHSI_framework}. The framework contains three main components: embedding layer, encoder backbone, segmentation head.

\textbf{Embedding layer} projects the spectral vector into an embedding space. Notably, different from the existing embedding methods based on patches, the embedding layer extracts embeddings of each pixel. This enables our model to obtain more fine-grained pixel embeddings, which is more suitable for dense prediction tasks such as hyperspectral classification.
\blue{Specifically, the fine-grained pixel embedding $\mathbf{E}\in R^{H \times W \times D}$ can be obtained from hyperspectral image $\mathbf{I} \in R^{H \times W\times C}$ as follows:}
\begin{equation}
    \begin{split}
    \mathbf{E} &= \rm{Embedding}(\mathbf{I}) \\
    &=\rm{SiLU}(\rm{GN}(\rm{Conv}(\mathbf{I}))),
    \end{split}
\end{equation}
\blue{where Conv, GN, and SiLU denote the convolutional layer with $1\times 1$ kernel size, the group norm layer, and SiLU activation function, respectively. $H$, $W$, and $C$ signify the height, width, spectral channel number of the input hyperspectral image, respectively. $D$ is the embedding dimension. $\mathbf{I}$ and $\mathbf{E}$ denote the input hyperspectral image and the extracted embedding.}

\textbf{Encoder backbone} is used to extract the discriminative spatial-spectral features for classification. Specifically, the encoder backbone mainly contains three components: spatial Mamba block for extracting spatial features, spectral Mamba block for capturing spectral features, spatial-spectral fusion module for integrating spatial and spectral features. The process can be defined as follows:
\begin{equation}
    \mathbf{H} = \rm{Encoder}(\mathbf{E}),
\end{equation}
where $\rm{Encoder}$ and $\mathbf{H}$ denote the encoder backbone and the extracted hidden features.

\textbf{Segmentation head} adopts convolutional layer with $1 \times 1$ kernel size to obtain the final logits $l$, \textit{i.e.}, $\mathbf{l}=\rm{SegHead}(\mathbf{H})$. 

\subsection{Spatial Mamba Block}
\label{sec:SpaMB}
The hyperspectral classification is a pixel-level classification task. This means that the representation used for classification needs to satisfy two conditions. First, the representation should be refined and should reflect the differences between pixels. Therefore, different from the existing patch methods, we extract embedding in a pixel-level manner. Second, the representations should be discriminative for classification. Hence, we require that the proposed modules have strong long-distance modeling capabilities. Notably, the transformer has quadratic computational complexity and cannot be used to establish long-distance dependencies at the pixel level. This compels us to design a novel spatial feature extractor to build long-range dependencies in linear computational complexity. In this paper, we adopt the Mamba layer as the basic unit to build the spatial feature extractor, which can model long-range dependencies in linear computational complexity.

The detailed structure of the spatial Mamba block (SpaMB) is shown in ~\cref{fig:MambaHSI_framework}. The forward process can be formulated as follows:
\begin{equation}
    \begin{split}
        & \mathbf{HF_{spa}} = \rm{Flatten}(\mathbf{H^{i}}), \\
        & \mathbf{HR_{spa}} = \rm{SiLU}(\rm{GN}(\rm{Mamba}(\mathbf{HF_{spa}}))), \\
        & \mathbf{H_{spa}^{o}} = \rm{Reshape}(\mathbf{HR_{spa}}) + \mathbf{H^{i}},
    \end{split}
\end{equation}
\blue{where $\mathbf{H^{i}} \in \mathcal{R}^{B \times H \times W \times D}$, $\mathbf{H_{spa}^{o}} \in \mathcal{R}^{B \times H \times W \times D}$ denote the input embeddings of pixel-level and the output features of SpaMB. $B,H,W,D$ signify the batch size, image height, image width, and embedding dimension, respectively. The embedding dimension $D$ is set to 128 in experiments. $\mathbf{HF_{spa}} \in \mathcal{R}^{B \times L1 \times D}$ and $\mathbf{HR_{spa}} \in \mathcal{R}^{B \times L1 \times D}$ denote the flatten input and the learned residual spatial feature. $L1$ is equal to $H \times W$. The Mamba denotes the standard Mamba block proposed in the~\cite{gu2023mamba}.
The design of group norm (GN) and residual connection aid the SpaMB learning}.

\subsection{Spectral Mamba Block}
Unlike traditional visual systems that capture image through RGB channels, hyperspectral images can cover a much larger spectral range and higher spectral resolution. How to model the relationship between spectra and extract discriminative features is still an open research problem. In this paper, we design a spectral Mamba block (SpeMB) to achieve the above goal. The overall structure is shown in ~\cref{fig:MambaHSI_framework}. Specifically, we divide a spectral features into $G$ group. Then we model the relations between different spectral groups, and then update the spectral features guided by the mined relationships between spectral groups. The extracted spectral features $H^{o}$ can be obtained as follows:
\begin{equation}
    \begin{split}
        & \mathbf{HG_{spe}} = \rm{SplitSpectralGroup}(\mathbf{H^{i}}), \\
        & \mathbf{HF_{spe}} = \rm{Flatten}(\mathbf{HG_{spe}}), \\
        & \mathbf{HR_{spe}} = \rm{SiLU}(\rm{GN}(\rm{Mamba}(\mathbf{HF_{spe}}))), \\
        & \mathbf{H_{spe}^{o}} = \rm{Reshape}(\mathbf{HR_{spe}}) + \mathbf{H^{i}},
    \end{split}
\end{equation}
where $\mathbf{HG_{spe}} \in \mathcal{R}^{B \times H \times W \times G \times M}$, $\mathbf{HF_{spe}} \in \mathcal{R}^{N \times G \times M}$, $\mathbf{HR_{spe}} \in \mathcal{R}^{N \times G \times M}$, and $\mathbf{H_{spe}^{o}} \in \mathcal{R}^{B \times H \times W \times D}$ denote the divided spectral group features, the flatten ones, the residual ones, and the output spectral features, respectively. $G$ represents the number of groups into which the semantic vector is split. $M$ equals the pixel embedding dimension $D$ divided by the group number $G$. $N$ is equal to $B \times H \times W$. \blue{The Mamba signifies the standard mamba block proposed in the~\cite{gu2023mamba}.}
\subsection{Spatial-Spectral Fusion Module}
Both spatial and spectral features are vital for HSI classification, and integrating spatial and spectral information is beneficial for classification~\cite{SSFCN}. This motivates us to design a spatial-spectral fusion module (SSFM), and the architechture of SSFM is shown in ~\cref{fig:MambaHSI_framework}. Considering that hyperspectral classification usually has fewer labeled samples, we introduced the idea of residual learning to alleviate overfitting that may occur in the training process. Besides, the SSFM adaptively estimates the importance of spatial and spectral to guide the fusion. The fusion process can be formulated as follows:
\begin{equation}
    \begin{split}
        \mathbf{H_{fus}} = \mathbf{H_{i}} + w_{spa} \times \mathbf{H_{spa}^{o}} + w_{spe} \times \mathbf{H_{spe}^{o}},
    \end{split}
\end{equation}
where $w_{spa}$ and $ w_{spe}$ denote the fusion weight of spatial and spectral, respectively. \blue{$w_{spa}$ and $ w_{spe}$ are randomly initialized, and these weights are updated through backpropagation to determine the final fusion weight.}

\blue{\subsection{Computational Complexity}}
\blue{Considering that the number of pixels L is much larger than the number of images B, the number of feature channels D and other parameters, we analyzed the relationship between the pixel number L and computational complexity. The computational complexity of a Transformer is quadratic with respect to the length of input sequence~\cite{transformer}, i.e., $O(L^2)$. In addition, Mamba has a linear complexity with respect to the sequence length~\cite{gu2023mamba}, i.e., $O(L)$. The proposed MambaHSI consists of embedding layer, spatial mamba block, spectral mamba block, spatial-spectral fusion module, and segmentation head. Their complexity all are $O(L)$. Therefore, the complexity of the proposed MambaHSI is $O(L)$.}

\subsection{Training and Inference}
\begin{algorithm}[!t]
	\SetAlgoLined
        \SetArgSty{textnormal}
        \LinesNumbered
        \textcolor[RGB]{61,102,150}{\small{\tcp{Training}}}
        \small{\KwIn{Whole image $\mathbf{I}$ and training label $\mathbf{Y_{tr}}$}}
        
       \textbf{Initialize: }Initialize network parameters $\Theta$ with random values\\

	\For{$e = 1, 2, \dots, E$}{
		\textcolor[RGB]{61,102,150}   {\small{\tcp{Whole Image Forward Process}}}
            
            \blue{$\mathbf{E} = \rm{Embedding}(\mathbf{I})$;} \small{\textcolor{gray}{//\blue{Extract Image Embedding}}}
            
            \blue{$\mathbf{H} = \rm{Encoder}(\mathbf{E})$;} \small{\textcolor{gray}{//\blue{Encoder incudes SpaMB, SpeMB, and SSFM modules.}}}

            \blue{$\mathbf{l} = \rm{SegHead}(\mathbf{H})$;} \small{\textcolor{gray}{//\blue{Obtain Image Logits}}}
            
            \textcolor[RGB]{61,102,150}   {\small{\tcp{Compute Loss}}}
            $\mathcal{L} \leftarrow \left\{\mathbf{l}, \mathbf{Y_{tr}^{s}}\right\}$ in Eq.(\ref{eq:loss}).
           
            \textcolor[RGB]{61,102,150}{\small{\tcp{Update Parameters}}}
            
            Update the parameters of MambaHSI using Adam.
            
	}
        \small{\KwOut{Network parameters $\Theta$}} 
        
        \textcolor[RGB]{61,102,150}{\small{\tcp{Testing}}}

        \small{\KwIn{Whole test image $\mathbf{I_{te}}$}}
        1) Load trained network parameter $\Theta$ \\
        2) predict the testing image with the trained model\\
        \KwOut{The class prediction $\hat{y}$}

        \caption{\small{Training and test procedures of MambaHSI}}
	\label{alg:algorithm}
\end{algorithm}

\blue{The training and test procedure is presented in Algorithm 1. Unlike patch-level HSI classification methods~\cite{CLOLN,GSC-Vit_TGRS24}, MambaHSI is an image-level HSI method, which adopts the whole image as input and can obtain all pixel’s prediction of whole image by one time forward. Additionally, the encoder, including the Spatial Mamba Block (SpaMB), Spectral Mamba Block (SpeMB), and Spatial-Spectral Fusion Module (SSFM), extracts the spatial-spectral features. MambaHSI can be trained in an end-to-end manner.}
In training phase, the training loss function $\mathcal{L}$ is given as:
\begin{equation}
    \label{eq:loss}
    \mathcal{L} = CrossEntropy(\mathbf{l}, \mathbf{Y_{tr}}),
\end{equation}
where $\mathbf{Y_{tr}}$ is the training label and $\mathbf{l}=\rm{MambaHSI}(\mathbf{I})$ is the logits output by MambaHSI. $\mathbf{I}$ is the whole image.

In inference phase, given a test image ($\mathbf{I_{te}}$), the final prediction $\hat{y}$ can be obtained as follows:
\begin{equation}
    \hat{y} = \arg \max(softmax(\rm{MambaHSI}(\mathbf{I_{te}}))).
\end{equation}

\section{Experiments}
In this section, we first describe the experimental setup, including datasets, evaluation metrics, comparison methods, and implementation details. Then we comprehensively compare with state-of-the-arts methods both qualitatively and quantitatively. Finally, we conduct a detailed ablation study to analyze the effect of the proposed modules.
\subsection{Experimental Setup}
\noindent \textbf{Datasets.} To more comprehensively evaluate the effectiveness of the proposed model, we choose four widely used and diverse hyperspectral datasets, \textit{i.e.}, Pavia University (PaviaU), Houston, WHU-Hi-HanChuan (HanChuan)~\cite{WHU-Hi} and WHU-Hi-HongHu (HongHu)~\cite{WHU-Hi}.
\subsubsection{Pavia University} 
\begin{figure}[!t]
    \noindent \begin{minipage}[]{\linewidth}
    \centering
    \subfloat[]{\includegraphics[width=0.40\linewidth]{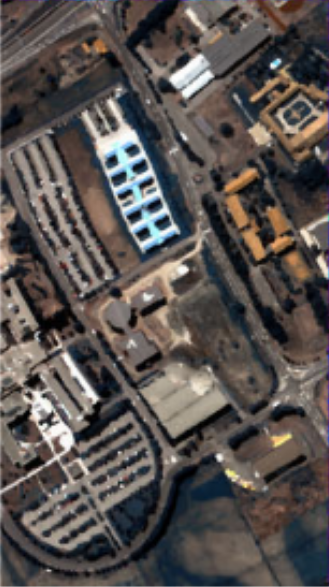}}
    \subfloat[]{\includegraphics[width=0.40\linewidth]{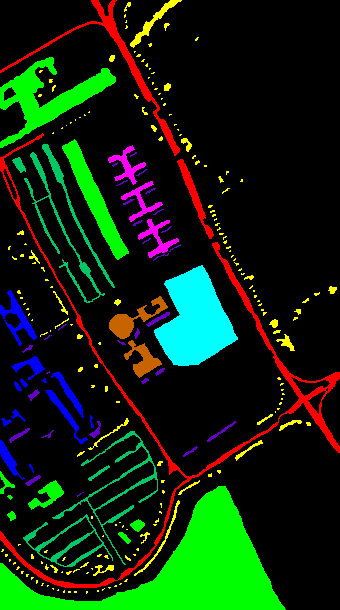}}
    \end{minipage}
    \hfill
    
    \noindent \begin{minipage}[]{\linewidth}
    \centering
    \small{
    \resizebox{\linewidth}{!}{
    \begin{tabular}{ccccccc}
        \hline
        ID & Color & Category & Train & validation & Test & Total \\ \hline
        1 & \cellcolor{c1} & Asphalt & 30 & 10 & 6591 & 6631 \\
        2 & \cellcolor{c2} & Meadows & 30 & 10 & 18609 & 18649 \\
        3 & \cellcolor{c3} & Gravel & 30 & 10 & 2059 & 2099 \\
        4 & \cellcolor{c4} & Trees & 30 & 10 & 3024 & 3064 \\
        5 & \cellcolor{c5} & Metal sheets & 30 & 10 & 1305 & 1345 \\
        6 & \cellcolor{c6} & Bare soil & 30 & 10 & 4989 & 5029 \\
        7 & \cellcolor{c7} & Bitumen & 30 & 10 & 1290 & 1330 \\
        8 & \cellcolor{c8} & Bricks & 30 & 10 & 3642 & 3682 \\
        9 & \cellcolor{c9} & Shadows & 30 & 10 & 907 & 947 \\ \hline
        Total & & & 270 & 90 & 42416 & 42776 \\ \hline
    \end{tabular}
    }}
    \put(-132,-60){(c)}
    \end{minipage}
    \vspace{-5pt}
    \caption{\small{Pavia Univeristy data set. (a) False color image. (b) Ground truth. (c) Category and sample settings.}}
    \vspace{-15pt}
    \label{fig:PaviaU_dataset}
\end{figure}

The image was captured by the Reflective Optics System Imaging Spectrometer (ROSIS) sensor over Pavia University, consisting $103$ bands with a spatial size of $610 \times 340$. The dataset contains $42776$ labeled pixels of nine classes. The related information, such as original image visualization, ground truth map, and the dataset split configuration, is shown in ~\cref{fig:PaviaU_dataset}.

\subsubsection{Houston} 
\begin{figure}[!t]
    \noindent \begin{minipage}[]{\linewidth}
    \centering
    \subfloat[]{\includegraphics[width=\linewidth]{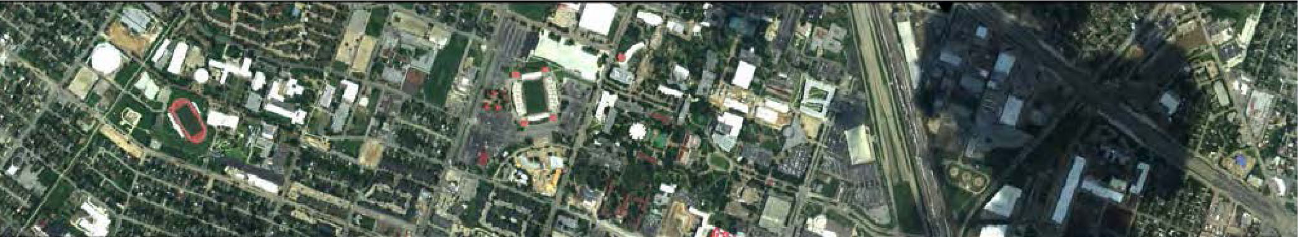}}
    \vspace{-1pt}
    \subfloat[]{\includegraphics[width=\linewidth]{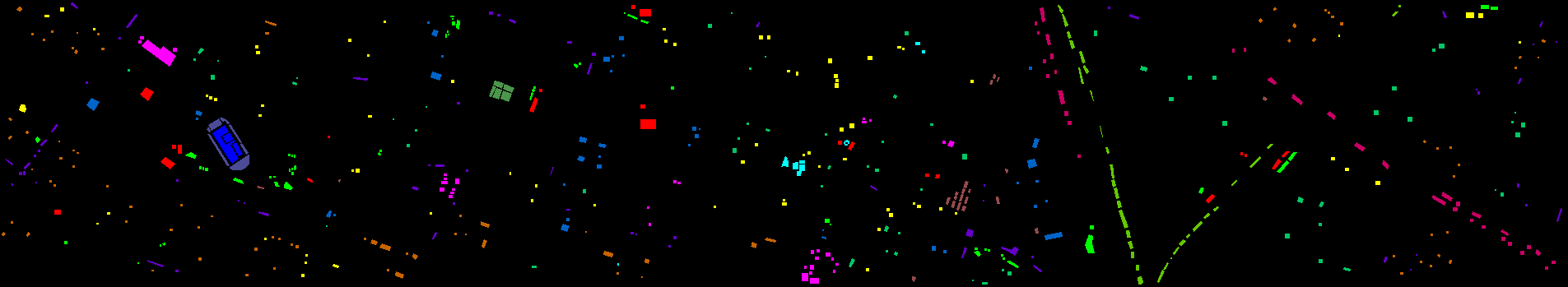}}
    \end{minipage}
    \vfill
    
    \noindent \begin{minipage}[]{\linewidth}
    \centering
    \small{
    \resizebox{\linewidth}{!}{
    \begin{tabular}{ccccccc}
        \hline
        ID & Color & Category & Train & validation & Test & Total \\ \hline
        1 & \cellcolor{c1} & Healthy Grass & 30 & 10 & 1211 & 1251 \\
        2 & \cellcolor{c2} & Stressed Grass & 30 & 10 & 1214 & 1254 \\
        3 & \cellcolor{c3} & Synthetic Grass & 30 & 10 & 657 & 697 \\
        4 & \cellcolor{c4} & Tree & 30 & 10 & 1204 & 1244 \\
        5 & \cellcolor{c5} & Soil & 30 & 10 & 1202 & 1242 \\
        6 & \cellcolor{c6} & Water & 30 & 10 & 285 & 325 \\
        7 & \cellcolor{c7} & Residential & 30 & 10 & 1228 & 1268 \\
        8 & \cellcolor{c8} & Commercial & 30 & 10 & 1204 & 1244 \\
        9 & \cellcolor{c9} & Road & 30 & 10 & 1212 & 1252 \\ 
        10 & \cellcolor{c10} & Highway & 30 & 10 & 1187 & 1227 \\
        11 & \cellcolor{c11} & Railway & 30 & 10 & 1195 & 1235 \\
        12 & \cellcolor{c12} & Parking Lot 1 & 30 & 10 & 1193 & 1233 \\
        13 & \cellcolor{c13} & Parking Lot 2 & 30 & 10 & 429 & 469 \\
        14 & \cellcolor{c14} & Tennis Court & 30 & 10 & 388 & 428 \\
        15 & \cellcolor{c15} & Running Track & 30 & 10 & 620 & 660 \\
        \hline
        Total & & & 450 & 150 & 14429 & 15029 \\ \hline
    \end{tabular}
    }}
    \put(-132,-85){(c)}
    \end{minipage}
    \vspace{-5pt}
    \caption{\small{Houston data set. (a) False color image. (b) Ground truth. (c) Category and sample settings.}}
    \vspace{-15pt}
    \label{fig:Houston_dataset}
\end{figure}

The image was acquired over the University of Houston campus and its neighbor regions by ITRES CASI 1500 HS imager with 144 spectral bands
, and has an image size of $349 \times 1905$. It was provided by the $2013$ IEEE Geoscience and Remote Sensing Society (GRSS) data fusion contest. The dataset contains $16$ classes, including residual, commercial, and so on. More details are shown in ~\cref{fig:Houston_dataset}.

\subsubsection{HanChuan} 
\begin{figure}[!t]
    \noindent \begin{minipage}[]{\linewidth}
    \centering
    \subfloat[]{\rotatebox{90}{\includegraphics[height=\linewidth]{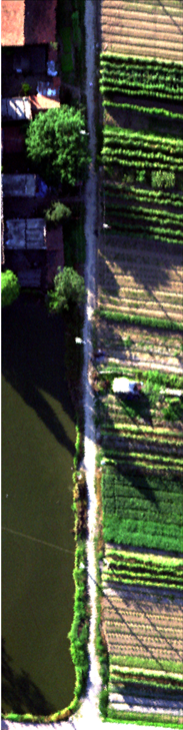}}}
    \vspace{-1pt}
    \subfloat[]{\rotatebox{90}{\includegraphics[height=\linewidth]{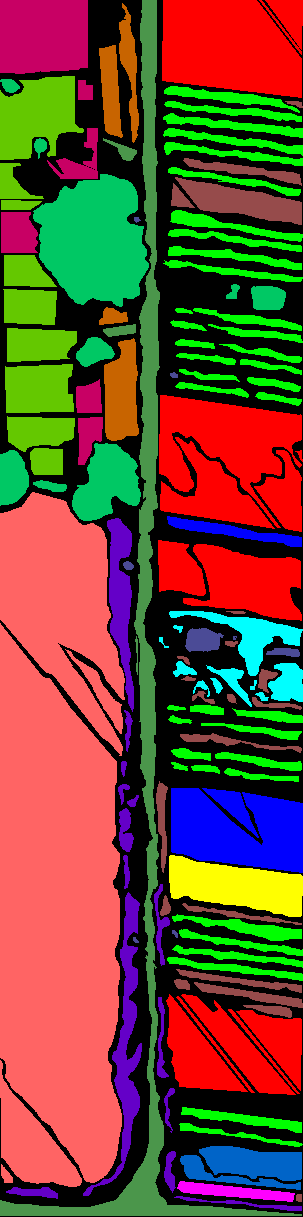}}}
    \end{minipage}
    \hfill
    
    \noindent \begin{minipage}[]{\linewidth}
    \centering
    \small{
    \resizebox{\linewidth}{!}{
    \begin{tabular}{ccccccc}
        \hline
        ID & Color & Category & Train & validation & Test & Total \\ \hline
        1 & \cellcolor{c1} & Strawberry & 30 & 10 & 44695 & 44735 \\
        2 & \cellcolor{c2} & Cowpea & 30 & 10 & 22713 & 22753 \\
        3 & \cellcolor{c3} & Soybean & 30 & 10 & 10247 & 10287 \\
        4 & \cellcolor{c4} & Sorghum & 30 & 10 & 5313 & 5353 \\
        5 & \cellcolor{c5} & Water spinach & 30 & 10 & 1160 & 1200 \\
        6 & \cellcolor{c6} & Watermelon & 30 & 10 & 4493 & 4533 \\
        7 & \cellcolor{c7} & Greens & 30 & 10 & 5863 & 5903 \\
        8 & \cellcolor{c8} & Trees & 30 & 10 & 17938 & 17978 \\
        9 & \cellcolor{c9} & Grass & 30 & 10 & 9429 & 9469 \\ 
        10 & \cellcolor{c10} & Red roof & 30 & 10 & 10476 & 10516 \\
        11 & \cellcolor{c11} & Gray roof & 30 & 10 & 16871 & 16911 \\
        12 & \cellcolor{c12} & Plastic & 30 & 10 & 3639 & 3679 \\
        13 & \cellcolor{c13} & Bare soil & 30 & 10 & 9076 & 9116 \\
        14 & \cellcolor{c14} & Road & 30 & 10 & 18520 & 18560 \\
        15 & \cellcolor{c15} & Bright object & 30 & 10 & 1096 & 1136 \\
        16 & \cellcolor{c16} & Water & 30 & 10 & 75361 & 75401 \\
        \hline
        Total & & & 480 & 160 & 256890 & 257530 \\ \hline
    \end{tabular}
    }}
    \put(-132,-85){(c)}
    \end{minipage}
    \vspace{-5pt}
    \caption{\small{HanChuan data set. (a) False color image. (b) Ground truth. (c) Category and sample settings.}}
    \vspace{-15pt}
    \label{fig:HanChuan_dataset}
\end{figure}

The WHU-Hi-HanChuan dataset was acquired from 17:57 to 18:46 on June $17$, $2016$, in Hanchuan, Hubei province, China, with an $17$ mm focal length Headwall Nano-Hyperspec imaging sensor equipped on a Leica Aibot X6 UAV V1 platform. The study area is a rural-urban fringe zone with buildings, water, and cultivated land, which contains seven crop species: strawberry, cowpea, soybean, sorghum, water spinach, watermelon, and greens. The size of the imagery is $1217 \times 303$ pixels, and there are $274$ bands from $400$ to $1000$ nm. An overview of this dataset is given in ~\cref{fig:HanChuan_dataset}.

\begin{figure}[!t]
    \noindent \begin{minipage}[]{\linewidth}
    \centering
    \subfloat[]
    {\includegraphics[width=0.35\linewidth]{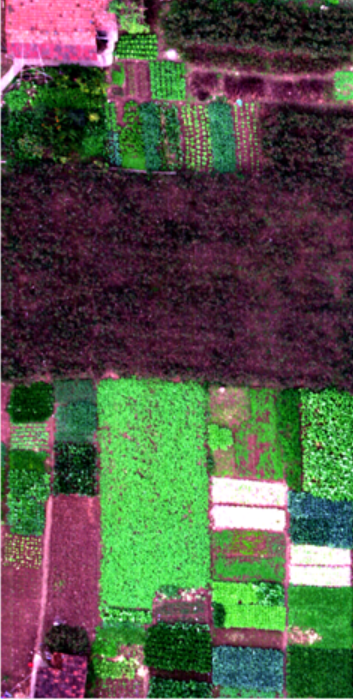}}
    \subfloat[]{\includegraphics[width=0.35\linewidth]{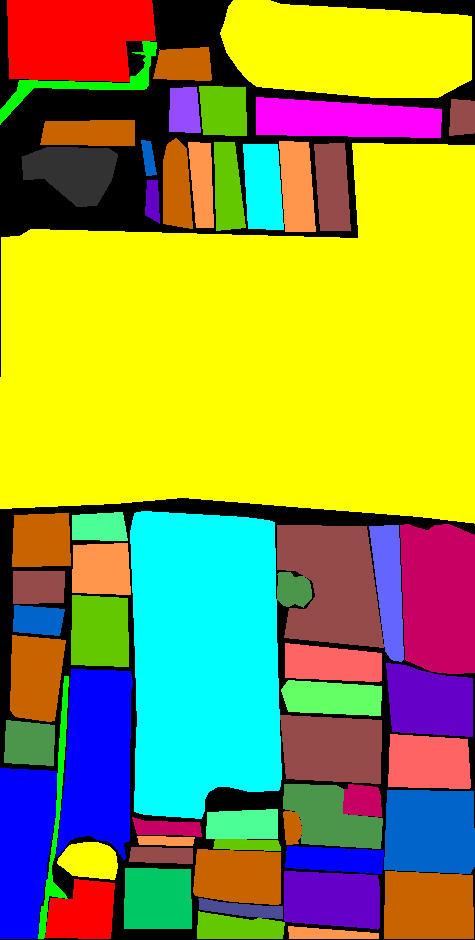}}
    \end{minipage}
    \hfill
    
    \noindent \begin{minipage}[]{\linewidth}
    \centering
    \small{
    \resizebox{\linewidth}{!}{
    \begin{tabular}{ccccccc}
        \hline
        ID & Color & Category & Train & validation & Test & Total \\ \hline
        1 & \cellcolor{c1} & Red roof & 30 & 10 & 14001 & 14041 \\
        2 & \cellcolor{c2} & Road & 30 & 10 & 3472 & 3512 \\
        3 & \cellcolor{c3} & Bare soil & 30 & 10 & 21781 & 21821 \\
        4 & \cellcolor{c4} & Cotton & 30 & 10 & 163245 & 163285 \\
        5 & \cellcolor{c5} & Cotton firewood & 30 & 10 & 6178 & 6218 \\
        6 & \cellcolor{c6} & Rape & 30 & 10 & 44517 & 44557 \\
        7 & \cellcolor{c7} & Chinese cabbage & 30 & 10 & 24063 & 24103 \\
        8 & \cellcolor{c8} & Pakchoi & 30 & 10 & 4014 & 4054 \\
        9 & \cellcolor{c9} & Cabbage & 30 & 10 & 10779 & 10819 \\ 
        10 & \cellcolor{c10} & Tuber mustard & 30 & 10 & 12354 & 12394 \\
        11 & \cellcolor{c11} & Brassica parachinensis	 & 30 & 10 & 10975 & 11015 \\
        12 & \cellcolor{c12} & Brassica chinensis & 30 & 10 & 8914 & 8954 \\
        13 & \cellcolor{c13} & Small Brassica chinensis & 30 & 10 & 22467 & 22507 \\
        14 & \cellcolor{c14} & Lactuca sativa & 30 & 10 & 7316 & 7356 \\
        15 & \cellcolor{c15} & Celtuce & 30 & 10 & 962 & 1002 \\
        16 & \cellcolor{c16} & Film covered lettuce & 30 & 10 & 7222 & 7262 \\
        17 & \cellcolor{c17} & Romaine lettuce & 30 & 10 & 2970 & 3010 \\
        18 & \cellcolor{c18} & Carrot & 30 & 10 & 3177 & 3217 \\
        19 & \cellcolor{c19} & White radish & 30 & 10 & 8672 & 8712 \\
        20 & \cellcolor{c20} & Garlic sprout	 & 30 & 10 & 3446 & 3486 \\
        21 & \cellcolor{c21} & Broad bean & 30 & 10 & 1288 & 1328 \\
        22 & \cellcolor{c22} & Tree & 30 & 10 & 4000 & 4040 \\
        \hline
        Total & & & 660 & 220 & 385813 & 386693 \\ \hline
    \end{tabular}
    }}
    \put(-132,-100){(c)}
    \end{minipage}
    \vspace{-5pt}
    \caption{\small{HongHu data set. (a) False color image. (b) Ground truth. (c) Category and sample settings.}}
    \vspace{-5pt}
    \label{fig:HongHu_dataset}
\end{figure}
\subsubsection{HongHu} The image was acquired from 16:23 to 17:37 on November $20$, $2017$, in Honghu City, Hubei province, China, with a $17$ mm focal length Headwall Nano-Hyperspec imaging sensor equipped on a DJI Matrice $600$ Pro UAV platform. The experimental area is a complex agricultural scene with many classes of crops, and different cultivars of the same crop are also planted in the region. The size of the imagery is $940 \times 475$ pixels, there are $270$ bands from $400$ to $1000$ nm. More information is given in ~\cref{fig:HongHu_dataset}.

\noindent \textbf{Evaluation Metrics.} Three evaluation metrics widely used in the hyperspectral classification were adopted in the experiments, including overall accuracy (OA), average accuracy (AA), and kappa coefficient ($\kappa$). To eliminate the deviation caused by randomly selecting training samples, the main comparative experiments and ablation studies were conducted ten times, while all other analytical experiments were performed five times. The larger the mean value and the smaller the standard deviation of these metrics, the better the performance of the model.

\noindent \textbf{Comparison Methods.} We compare relative hyperspectral image classification methods: 

\noindent {- For ML based:}
\begin{itemize}
    \item \blue{SVM\pub{TGRS2004}~\cite{SVM_Comp}: The model adopts a support vector machine (SVM) to address hyperspectral image classification.}
    \item \blue{RF\pub{TGRS2005}~\cite{RF_Comp}: The model exploits random forest (RF) to classify hyperspectral image data.}
\end{itemize}
\noindent {- For GCN based}:
\begin{itemize}
    \item \blue{DMSGer\pub{TNNLS2022}~\cite{DMSGer}: The model utilizes a dynamic multi-scale GCN classifier to capture pixel-level and region-level features.}
    \item \blue{GiGCN\pub{TNNLS2022}~\cite{GiGCN}: The model exploits information both inside and outside superpixels to boost classification performance.}
\end{itemize}
\noindent {- For CNN based:}
\begin{itemize}
    \item FullyContNet\pub{TGRS2022}~\cite{FullyContNet}: The model combines a scale attention mechanism and a multi-scale feature extractor, such as paramid pooling, to obtain more effective contexts for classification. The input of this method is the original hyperspectral image.
    \item CLOLN\pub{TGRS2024}~\cite{CLOLN}: The model is a channel-layeroriented lightweight network to alleviate the number of parameters and computational complexity associated with CNNs.
\end{itemize}
\noindent {- For Transformer based:}
\begin{itemize}
    \item Spectralformer~\pub{TGRS2021}~\cite{spectralformer}: The model extracts spectrally local sequence information from neighboring bands of HS images, yielding group-wise spectral embeddings. 
    \item GSC-ViT\pub{TGRS2024}~\cite{GSC-Vit_TGRS24}: The model adopts a groupwise separable convolution ViT to capture local and global spectral-spatial information for HSI classification.
\end{itemize}

\noindent \textbf{Implement Details.} \blue{The proposed MambaHSI is implemented with pytorch. All experiments are randomly conducted ten times to eliminate the influence of random sample selection and parameter initialization. In each trial, the training and validation set are composed of $30$ and $10$ samples randomly selected from the images. The testing set is made up of the remaining samples. As the proposed MambaHSI takes the whole image as input and obtains the predictions at once, the batch size is set to $1$. The Adam optimizer is used to train the model. The learning rate is set to $0.0003$. The group number $G$ and hidden dimension $D$ are set to $4$ and $128$. The timescale parameter $\Delta$, projection parameters $B$ and $C$ in the mamba block follow the parameter settings of the Mamba~\cite{gu2023mamba}. All runs are conducted on the same hardware: NVIDIA GeForce RTX $3090$ GPU, $\times 64$ Intel Xeon Gold $6226R$ CPU and $256$ GB RAM.}

\begin{table*}[]
\begin{tabular}{l|cc|cc|cc|cc|c}
\hline
\rowcolor[HTML]{EFEFEF} 
 & \multicolumn{2}{c|}{ML-based}           & \multicolumn{2}{c|}{GCN-based}          & \multicolumn{2}{c|}{CNN-based}          & \multicolumn{2}{c|}{Transformer-based} & SSM-based                 \\ \cline{2-10} 
\rowcolor[HTML]{EFEFEF} 
 & SVM                 & RF               & DMSGer              & GiGCN            & FullyContNet     & CLOLN               & Spectralformer    & GSC-ViT           & MambaHSI                  \\ 
 \rowcolor[HTML]{EFEFEF} 
\multirow{-3}{*}{Class} & \tiny{TGRS2004} & \tiny{TGRS2005} & \tiny{TNNLS2022} & \tiny{TNNLS}                                 & \tiny{TGRS2022} & \tiny{TGRS2024} & \tiny{TGRS2021} & \tiny{TGRS2024}                                 & \tiny{Ours}  \\ \hline\hline 
 
Asphalt                     & 93.69±2.34          & 93.74±1.91       & 81.49±4.78          & 90.36±1.73       & 90.98±7.20       & \textbf{98.27±1.00} & 92.60±2.29        & {\ul 97.10±1.64}  & 94.35±1.73                \\
Meadows                     & 88.56±2.41          & 87.29±1.93       & 87.60±3.00          & {\ul 98.86±0.78} & 89.87±13.81      & \textbf{98.98±0.76} & 93.58±2.14        & 97.68±1.32        & 96.43±1.92                \\
Gravel                      & 50.42±5.13          & 48.47±4.74       & \textbf{97.15±1.28} & 91.66±6.33       & 91.75±5.61       & 87.83±7.54          & 60.90±8.15        & 86.46±7.42        & {\ul 93.39±5.35}          \\
Trees                       & 55.88±5.67          & 55.21±4.85       & 87.68±2.85          & 81.34±10.11      & {\ul 93.50±6.74} & \textbf{97.80±1.46} & 92.28±6.23        & 91.83±9.94        & 89.75±3.60                \\
Metal sheets                & 95.67±1.95          & 92.06±1.78       & {\ul 99.88±1.68}    & 91.15±1.31       & 99.54±0.60       & 99.68±0.59          & 98.00±1.72        & 99.45±0.53        & \textbf{99.99±0.02}       \\
Bare soil                   & 49.11±8.63          & 44.29±5.43       & \textbf{99.66±6.60} & {\ul 98.87±1.90} & 92.36±9.33       & 85.27±7.50          & 63.92±12.41       & 79.65±9.24        & 98.60±0.93                \\
Bitumen                     & 41.46±3.09          & 48.42±4.67       & \textbf{99.36±0.88} & {\ul 99.28±0.61} & 96.34±6.58       & 85.92±7.00          & 48.11±8.47        & 86.63±10.10       & 96.50±3.13                \\
Bricks                      & 73.17±4.61          & 70.94±6.73       & 88.65±2.94          & 87.68±7.46       & 89.67±4.82       & {\ul 90.93±4.12}    & 81.22±4.66        & 89.50±4.70        & \textbf{94.40±2.93}       \\
Shadows                     & \textbf{99.91±0.08} & {\ul 99.81±0.11} & 96.89±0.82          & 97.21±3.36       & 96.73±9.02       & 97.12±2.03          & 88.78±5.92        & 98.71±1.10        & 99.36±0.62                \\ \hline
OA(\%)                      & 71.31±3.15          & 70.68±2.61       & 89.55±1.22          & 94.47±0.97       & 91.31±9.36       & {\ul 94.97±1.43}    & 91.10±2.60        & 92.65±2.30        & \textbf{95.74±0.90}       \\
AA(\%)                      & 71.98±1.39          & 71.14±1.61       & 93.15±0.70          & 92.93±1.07       & 93.41±6.57       & {\ul 93.53±1.59}    & 79.93±2.67        & 91.89±1.73        & \textbf{95.86±1.11}       \\
Kappa(\%)                                         & 63.96±3.19              & 63.13±2.81             & 86.49±1.54                 & 92.67±1.26                & 93.12±5.19       & {\ul 93.38±1.84}    & 77.40±4.41       & 90.38±2.95         & \textbf{95.00±2.24}           \\ \hline
\end{tabular}
\vspace{-5pt}
\caption{\small{Quantitative result (ACC\%±STD\%) of Pavia University. Best in \textbf{bold} and second with \uline{underline}. These notes are the same to others.}}
\vspace{-15pt}
\label{tab:comp_PaviaU}
\end{table*}

\begin{table*}[]
\setlength{\tabcolsep}{1.8mm}
\begin{tabular}{l|cc|cc|cc|cc|c}
\hline
\rowcolor[HTML]{EFEFEF} 
 & \multicolumn{2}{c|}{ML-based}           & \multicolumn{2}{c|}{GCN-based}          & \multicolumn{2}{c|}{CNN-based}          & \multicolumn{2}{c|}{Transformer-based} & SSM-based                 \\ \cline{2-10} 
\rowcolor[HTML]{EFEFEF} 
 & SVM                 & RF               & DMSGer              & GiGCN            & FullyContNet     & CLOLN               & Spectralformer    & GSC-ViT           & MambaHSI                  \\ 
 \rowcolor[HTML]{EFEFEF} 
\multirow{-3}{*}{Class} & \tiny{TGRS2004} & \tiny{TGRS2005} & \tiny{TNNLS2022} & \tiny{TNNLS2022}                                 & \tiny{TGRS2022} & \tiny{TGRS2024} & \tiny{TGRS2021} & \tiny{TGRS2024}                                 & \tiny{Ours}  \\ \hline\hline 
Healthy Grass                                     & {\ul 93.68±5.88}        & 93.00±0.45             & 91.27±4.59                 & 83.21±5.03                & 92.69±7.57           & 88.24±9.00          & 91.29±5.38      & 90.16±6.48          & {\ul \textbf{95.20±4.66}}     \\
Stressed Grass                                    & 85.92±6.74              & 91.05±2.57             & 91.29±4.73                 & 82.83±7.41                & 85.06±16.86          & 93.41±5.99          & 92.12±6.99      & {\ul 95.77±2.95}    & \textbf{98.29±1.26}           \\
Synthetic Grass                                   & 98.83±1.38              & 97.82±3.65             & 99.62±0.39                 & 98.83±1.66                & 99.21±0.74           & 95.21±9.84          & 94.68±3.25      & \textbf{99.83±0.26} & {\ul 99.74±0.46}              \\
Tree                                              & \textbf{98.89±0.50}     & 96.88±3.62             & 83.67±4.98                 & 72.42±10.94               & 93.11±3.19           & {\ul 97.44±3.80}    & 93.32±9.22      & 94.78±7.69          & 97.26±2.47                    \\
Soil                                              & 90.87±2.41              & 87.75±4.04             & {\ul 98.76±1.79}           & 93.14±4.62                & 97.07±6.99           & 97.07±2.86          & 98.18±1.72      & 97.63±2.69          & \textbf{99.54±0.76}           \\
Water                                             & 87.86±14.30             & 87.30±0.68             & \textbf{99.79±0.63}        & 88.99±6.78                & {\ul 97.58±1.78}     & 94.85±4.72          & 87.21±8.11      & 96.03±2.59          & 97.47±2.33                    \\
Residential                                       & 73.99±4.46              & 74.97±4.97             & 85.21±3.03                 & 82.34±5.24                & 88.62±2.41           & \textbf{94.02±2.94} & 91.47±3.54      & {\ul 93.11±2.39}    & 93.03±2.25                    \\
Commercial                                        & 79.40±11.34             & 83.26±3.92             & 67.40±5.02                 & {\ul 91.02±6.25}          & 76.74±9.92           & \textbf{93.80±3.94} & 84.41±7.91      & 90.53±8.57          & 81.35±3.98                    \\
Road                                              & 65.11±1.49              & 68.92±3.43             & 84.79±3.03                 & 85.51±6.43                & 83.11±5.86           & {\ul 88.32±2.83}    & 87.75±3.83      & 85.20±4.96          & \textbf{90.03±3.40}           \\
Highway                                           & 70.50±4.76              & 73.48±6.01             & \textbf{98.16±1.89}        & 94.92±2.69                & 91.88±9.34           & 76.55±6.96          & 82.09±5.12      & 85.28±4.22          & {\ul 96.55±1.49}              \\
Railway                                           & 70.15±2.45              & 66.13±3.26             & {\ul 94.22±4.37}           & \textbf{96.52±3.78}       & 90.33±5.02           & 91.57±4.40          & 85.90±4.63      & 93.08±3.33          & 92.74±2.18                    \\
Parking Lot 1                                     & 64.23±4.99              & 58.37±5.03             & {\ul 91.48±4.79}           & \textbf{92.45±3.63}       & 83.43±9.23           & 83.37±6.17          & 85.33±5.87      & 88.46±3.14          & 91.19±3.68                    \\
Parking Lot 2                                     & 43.46±10.62             & 31.71±4.80             & {\ul 96.95±2.60}           & 88.10±5.13                & 92.98±4.32           & 89.11±4.64          & 86.81±6.28      & 90.01±4.69          & \textbf{97.67±2.04}           \\
Tennis Court                                      & 88.44±5.61              & 80.89±5.68             & \textbf{100.00±0.00}       & 98.30±2.67                & \textbf{100.00±0.00} & 97.98±2.63          & 91.65±6.36      & 97.01±5.11          & \textbf{100.00±0.00}          \\
Running Track                                     & 99.16±0.46              & 97.08±1.84             & {\ul 99.97±0.10}           & 99.14±1.89                & 98.47±3.70           & 96.08±3.96          & 96.18±2.66      & 97.73±1.23          & \textbf{100.00±0.00}          \\ \hline
OA(\%)                                            & 79.63±1.04              & 78.95±0.90             & 90.36±1.29                 & 88.13±2.26                & 89.79±4.20           & 90.31±1.73          & 89.13±2.12      & {\ul 91.85±1.05}    & \textbf{94.46±0.83}           \\
AA(\%)                                            & 80.70±1.03              & 79.24±0.83             & 92.17±1.07                 & 89.85±1.91                & 91.35±3.55           & 91.80±1.46          & 89.89±1.74      & {\ul 92.97±0.92}    & \textbf{95.34±0.78}           \\
Kappa(\%)                                         & 77.97±1.12              & 77.25±0.98             & 89.58±1.39                 & 87.16±2.44                & 90.64±4.68           & 89.52±1.87          & 88.24±2.29      & {\ul 91.18±1.14}    & \textbf{94.21±1.93} \\ \hline         
\end{tabular}
\vspace{-5pt}
\caption{\small{Quantitative result (ACC\%±STD\%) of Houston dataset.}}
\vspace{-15pt}
\label{tab:comp_Houston}
\end{table*}

\begin{table*}[]
\setlength{\tabcolsep}{1.8mm}
\begin{tabular}{l|cc|cc|cc|cc|c}
\hline
\rowcolor[HTML]{EFEFEF} 
 & \multicolumn{2}{c|}{ML-based}           & \multicolumn{2}{c|}{GCN-based}          & \multicolumn{2}{c|}{CNN-based}          & \multicolumn{2}{c|}{Transformer-based} & SSM-based                 \\ \cline{2-10} 
\rowcolor[HTML]{EFEFEF} 
 & SVM                 & RF               & DMSGer              & GiGCN            & FullyContNet     & CLOLN               & Spectralformer    & GSC-ViT           & MambaHSI                  \\ 
 \rowcolor[HTML]{EFEFEF} 
\multirow{-3}{*}{Class} & \tiny{TGRS2004} & \tiny{TGRS2005} & \tiny{TNNLS2022} & \tiny{TNNLS2022}                                 & \tiny{TGRS2022} & \tiny{TGRS2024} & \tiny{TGRS2021} & \tiny{TGRS2024}                                 & \tiny{Ours}  \\ \hline\hline 
Strawberry                                        & 92.03±1.24              & 87.87±1.73             & 74.51±5.53                 & 89.34±2.23                & 78.59±12.03         & \textbf{95.97±1.87} & 95.13±1.25      & {\ul 95.26±1.50}    & 89.72±5.75                    \\
Cowpea                                            & 69.50±7.79              & 57.02±4.42             & 62.51±8.18                 & 81.64±3.38                & 84.21±4.34          & \textbf{92.48±2.56} & 87.27±3.37      & {\ul 91.22±3.05}    & 81.37±5.69                    \\
Soybean                                           & 47.41±4.94              & 45.31±2.24             & {\ul 94.63±2.80}           & 81.06±6.56                & 79.85±9.55          & 80.03±9.18          & 70.36±6.15      & 78.37±7.47          & \textbf{94.88±3.73}           \\
Sorghum                                           & 74.49±9.57              & 51.50±2.59             & \textbf{99.51±0.50}        & 67.09±4.82                & 84.68±13.06         & 91.12±7.72          & 88.62±6.67      & 85.17±12.38         & {\ul 98.46±1.35}              \\
Water spinach                                     & 9.72±0.77               & 9.25±0.78              & {\ul 99.18±0.92}           & 34.98±2.76                & 68.38±27.44         & 56.82±12.89         & 42.01±11.48     & 46.57±13.03         & \textbf{99.54±0.85}           \\
Watermelon                                        & 15.28±2.00              & 11.66±1.03             & \textbf{94.18±2.43}        & 53.81±9.81                & 9.39±15.90          & 38.07±8.70          & 34.68±6.40      & 35.97±7.01          & \textbf{76.99±4.95}           \\
Greens                                            & 36.69±2.30              & 45.58±3.46             & \textbf{98.72±1.26}        & 80.84±8.87                & {\ul 92.94±4.47}    & 68.87±17.46         & 67.69±10.29     & 66.90±15.67         & 92.61±4.63                    \\
Trees                                             & 65.71±5.51              & 66.22±4.84             & 84.57±4.09                 & {\ul 89.27±4.80}          & \textbf{91.24±1.96} & 81.52±7.87          & 83.17±4.16      & 82.39±5.23          & 70.73±7.10                    \\
Grass                                             & 47.93±7.71              & 40.87±7.40             & 72.25±5.64                 & 65.14±6.79                & {\ul 70.48±8.43}    & 68.59±6.61          & 63.00±4.53      & 60.64±17.62         & \textbf{89.01±3.62}           \\
Red roof                                          & 74.50±11.85             & 59.59±6.64             & 87.81±3.96                 & 93.03±4.21                & 85.23±6.83          & 90.79±14.71         & 88.76±12.99     & {\ul 93.22±4.31}    & \textbf{94.44±2.18}           \\
Gray roof                                         & 35.90±1.90              & 54.20±9.34             & \textbf{96.13±1.66}        & {\ul 93.86±5.06}          & 92.58±4.29          & 87.86±4.63          & 77.41±13.19     & 82.13±7.96          & 89.56±4.91                    \\
Plastic                                           & 13.49±2.57              & 17.50±2.66             & \textbf{99.09±1.04}        & {\ul 89.22±7.45}          & 39.55±27.62         & 63.48±14.80         & 51.40±6.89      & 51.92±6.01          & 88.83±6.91                    \\
Bare soil                                         & 27.36±3.66              & 33.89±4.49             & \textbf{82.67±3.97}        & 66.66±6.60                & {\ul 82.37±2.46}    & 50.21±6.71          & 47.62±9.17      & 53.28±10.88         & 78.69±3.73                    \\
Road                                              & 80.55±8.56              & 76.20±6.53             & 75.85±3.16                 & \textbf{91.25±2.43}       & 88.31±4.97          & {\ul 90.78±2.74}    & 88.92±3.08      & 90.64±3.07          & 89.68±3.03                    \\
Bright object                                     & 33.15±10.97             & 26.80±10.82            & \textbf{98.36±2.21}        & 46.57±17.42               & {\ul 97.30±2.18}    & 62.42±10.67         & 52.51±8.59      & 70.05±12.35         & 94.42±2.41                    \\
Water                                             & 99.95±0.06              & 98.82±1.33             & 92.23±5.31                 & {\ul 99.75±0.14}          & 93.64±4.75          & 98.97±1.02          & 99.47±0.33      & \textbf{99.77±0.23} & 98.28±1.62                    \\ \hline
OA(\%)                                            & 60.98±1.35              & 64.09±1.78             & 84.39±1.83                 & {\ul 86.78±1.43}          & 84.88±3.02          & 85.14±2.93          & 82.99±2.31      & 83.14±3.96          & \textbf{90.21±1.67}           \\
AA(\%)                                            & 51.48±1.46              & 48.89±1.15             & {\ul 88.26±0.89}           & 76.47±1.62                & 77.42±3.61          & 76.13±3.26          & 71.13±2.56      & 73.97±2.06          & \textbf{89.20±1.36}           \\
Kappa(\%)                                         & 56.00±1.49              & 59.05±1.86             & 82.01±2.02                 & {\ul 84.64±1.62}          & 75.80±4.70          & 82.75±3.31          & 80.29±2.61      & 80.57±4.33          & \textbf{89.07±2.31}          \\ \hline
\end{tabular}
\vspace{-5pt}
\caption{\small{Quantitative result (ACC\%±STD\%) of HanChuan dataset.}}
\vspace{-8pt}
\label{tab:comp_HanChuan}
\end{table*}

\begin{table*}[]
\setlength{\tabcolsep}{1.05mm}
\begin{tabular}{l|cc|cc|cc|cc|c}
\hline
\rowcolor[HTML]{EFEFEF} 
 & \multicolumn{2}{c|}{ML-based}           & \multicolumn{2}{c|}{GCN-based}          & \multicolumn{2}{c|}{CNN-based}          & \multicolumn{2}{c|}{Transformer-based} & SSM-based                 \\ \cline{2-10} 
\rowcolor[HTML]{EFEFEF} 
 & SVM                 & RF               & DMSGer              & GiGCN            & FullyContNet     & CLOLN               & Spectralformer    & GSC-ViT           & MambaHSI                  \\ 
 \rowcolor[HTML]{EFEFEF} 
\multirow{-3}{*}{Class} & \tiny{TGRS2004} & \tiny{TGRS2005} & \tiny{TNNLS2022} & \tiny{TNNLS2022}                                 & \tiny{TGRS2022} & \tiny{TGRS2024} & \tiny{TGRS2021} & \tiny{TGRS2024}                                 & \tiny{Ours}  \\ \hline\hline 
Red roof                                          & 94.30±1.90              & 84.35±5.74             & 92.96±4.02                 & 90.93±2.64                & {\ul 96.78±1.57}    & 95.36±5.56          & 96.66±1.63          & \textbf{98.16±1.08} & 96.47±1.50                    \\
Road                                              & 53.77±4.54              & 52.01±5.17             & 93.34±4.05                 & 51.88±7.77                & \textbf{95.63±2.54} & 78.60±6.19          & 62.84±4.24          & 73.78±9.09          & {\ul 95.43±2.33}              \\
Bare soil                                         & 90.70±3.65              & 90.92±2.70             & 92.15±3.46                 & 78.80±2.87                & 92.23±1.51          & \textbf{98.22±1.32} & 95.60±2.81          & 97.21±1.81          & 94.66±1.05                    \\
Cotton                                            & 97.42±0.34              & 97.09±0.51             & 60.26±10.53                & 99.37±0.32                & 93.59±2.92          & {\ul 99.48±0.37}    & 99.42±0.30          & \textbf{99.57±0.14} & 96.96±1.79                    \\
Cotton firewood                                   & 17.61±2.78              & 14.17±2.64             & \textbf{99.30±0.79}        & 62.41±8.43                & 96.03±2.60          & 41.33±18.92         & 42.14±16.25         & 37.03±9.33          & {\ul 96.20±1.77}              \\
Rape                                              & 90.59±1.93              & 84.04±1.60             & 84.36±4.27                 & 89.61±3.17                & 91.61±2.28          & \textbf{96.09±1.67} & {\ul 95.73±1.02}    & 95.72±1.83          & 93.90±2.47                    \\
Chinese cabbage                                   & 78.60±3.21              & 72.03±4.93             & 69.44±3.44                 & 82.91±3.07                & 82.52±3.43          & \textbf{92.72±1.96} & 90.05±2.26          & {\ul 90.38±3.66}    & 86.60±3.30                    \\
Pakchoi                                           & 13.56±2.31              & 9.13±2.07              & \textbf{98.36±2.87}        & 75.44±8.66                & {\ul 96.81±2.03}    & 46.44±6.02          & 35.59±5.04          & 32.12±4.83          & 96.08±2.90                    \\
Cabbage                                           & 96.57±1.71              & 90.60±3.89             & 95.14±1.73                 & 85.39±3.04                & 96.62±1.20          & 97.04±2.05          & \textbf{97.89±1.07} & {\ul 97.06±2.73}    & 94.78±1.69                    \\
Tuber mustard                                     & 50.25±7.08              & 33.29±6.19             & 87.22±4.16                 & \textbf{92.42±2.80}       & 90.08±2.81          & 84.88±3.47          & 78.81±4.58          & 79.23±7.53          & {\ul 90.45±2.72}              \\
Brassica parachinensis                            & 31.70±5.48              & 25.71±4.48             & 82.08±7.09                 & 75.17±4.50                & {\ul 86.27±3.66}    & 72.03±8.35          & 69.29±7.09          & 72.41±8.95          & \textbf{91.96±2.87}           \\
Brassica chinensis                                & 42.58±4.37              & 40.77±3.36             & \textbf{91.91±2.46}        & 82.52±5.36                & {\ul 88.62±4.09}    & 66.63±8.82          & 58.68±5.96          & 65.61±8.66          & 87.40±5.22                    \\
Small Brassica chinensis                          & 59.05±4.52              & 54.24±3.33             & 73.10±6.40                 & \textbf{89.28±2.70}       & {\ul 89.16±2.24}    & 82.67±5.83          & 76.20±6.07          & 77.49±8.47          & 89.12±3.12                    \\
Lactuca sativa                                    & 67.70±6.77              & 57.66±7.93             & \textbf{94.87±2.43}        & 68.03±5.44                & 93.88±1.93          & 87.59±7.03          & 83.89±6.15          & 86.26±11.93         & {\ul 94.01±1.65}              \\
Celtuce                                           & 8.86±4.25               & 3.78±1.16              & \textbf{99.85±0.28}        & 21.65±2.65                & 98.42±1.16          & 75.88±12.93         & 61.24±13.37         & 63.62±20.25         & {\ul 98.84±1.17}              \\
Film covered lettuce                              & 87.85±2.55              & 80.73±7.57             & 96.89±1.26                 & 71.70±6.79                & 88.96±4.28          & {\ul 97.39±1.73}    & 95.23±2.99          & \textbf{97.42±1.66} & 96.83±1.60                    \\
Romaine lettuce                                   & 55.60±4.87              & 54.14±6.89             & \textbf{99.23±1.01}        & 54.94±4.04                & 95.10±4.52          & 80.09±9.70          & 68.68±8.79          & 84.92±6.09          & {\ul 98.60±1.49}              \\
Carrot                                            & 28.46±3.59              & 18.32±1.20             & \textbf{98.44±2.49}        & 67.65±8.17                & {\ul 97.22±1.43}    & 65.55±15.89         & 60.99±7.01          & 67.44±10.50         & 96.72±1.82                    \\
White radish                                      & 65.15±5.12              & 47.85±5.57             & 88.34±4.02                 & 72.71±8.86                & {\ul 89.35±3.28}    & 82.53±7.23          & 82.77±4.78          & 85.10±8.11          & \textbf{91.30±3.07}           \\
Garlic sprout                                     & 37.85±5.53              & 17.29±4.70             & {\ul 99.02±0.85}           & 66.48±6.47                & 98.34±1.10          & 81.38±6.60          & 71.18±8.29          & 71.64±10.46         & \textbf{99.18±0.42}           \\
Broad bean                                        & 9.35±0.92               & 9.68±0.83              & \textbf{99.96±0.12}        & 65.20±14.58               & {\ul 99.95±0.12}    & 30.19±12.06         & 28.91±8.18          & 25.92±7.91          & 99.91±0.18                    \\
Tree                                              & 21.53±1.27              & 20.00±1.27             & \textbf{99.88±0.24}        & 85.35±8.84                & 99.11±1.00          & 55.02±10.89         & 58.01±15.28         & 46.26±11.15         & {\ul 99.54±0.51}              \\ \hline
OA(\%)                                            & 68.77±1.27              & 59.85±2.13             & 75.43±4.34                 & 87.22±0.60                & {\ul 92.19±1.35}    & 87.93±2.11          & 86.04±2.83          & 86.18±1.86          & \textbf{94.58±1.01}           \\
AA(\%)                                            & 54.50±0.41              & 48.08±1.05             & 90.73±0.92                 & 74.08±1.50                & {\ul 93.47±0.48}    & 77.60±2.25          & 73.17±2.45          & 74.74±1.64          & \textbf{94.77±0.61}           \\
Kappa(\%)                                         & 62.71±1.28              & 53.22±1.95             & 71.59±4.32                 & 83.95±0.73                & {\ul 93.86±1.54}    & 85.06±2.49          & 82.72±3.22          & 82.93±2.16          & \textbf{94.00±1.88}    \\ \hline      
\end{tabular}
\vspace{-5pt}
\caption{\small{Quantitative result (ACC\%±STD\%) of HongHu dataset.}}
\vspace{-15pt}
\label{tab:comp_HongHu}
\end{table*}

\begin{figure*}
    \centering
    \includegraphics[width=\linewidth]{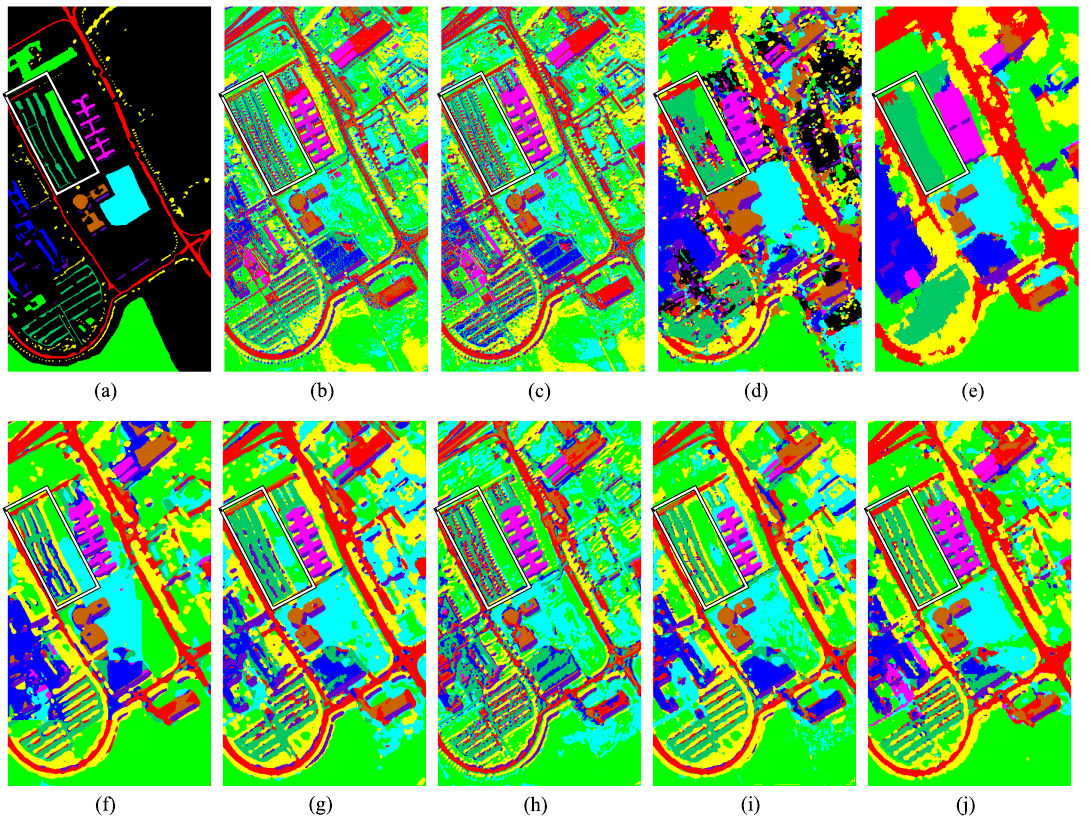}
    \vspace{-5pt}
    \caption{\small{Qualitative visualization of the classification map for Pavia University dataset. (a) Ground-truth map. (b) SVM. (c) RF. (d) DMSGer. (e) GiGCN. (f) FullyContNet. (g) CLOLN. (h) Spectralformer. (i) GSC-ViT. (j) The proposed MambaHSI. The meaning of colors refers to ~\cref{fig:PaviaU_dataset} c.}}
    \vspace{-15pt}
    \label{fig:vis_PaviaU}
\end{figure*}

\begin{figure*}
    \centering
    \includegraphics[width=\linewidth]{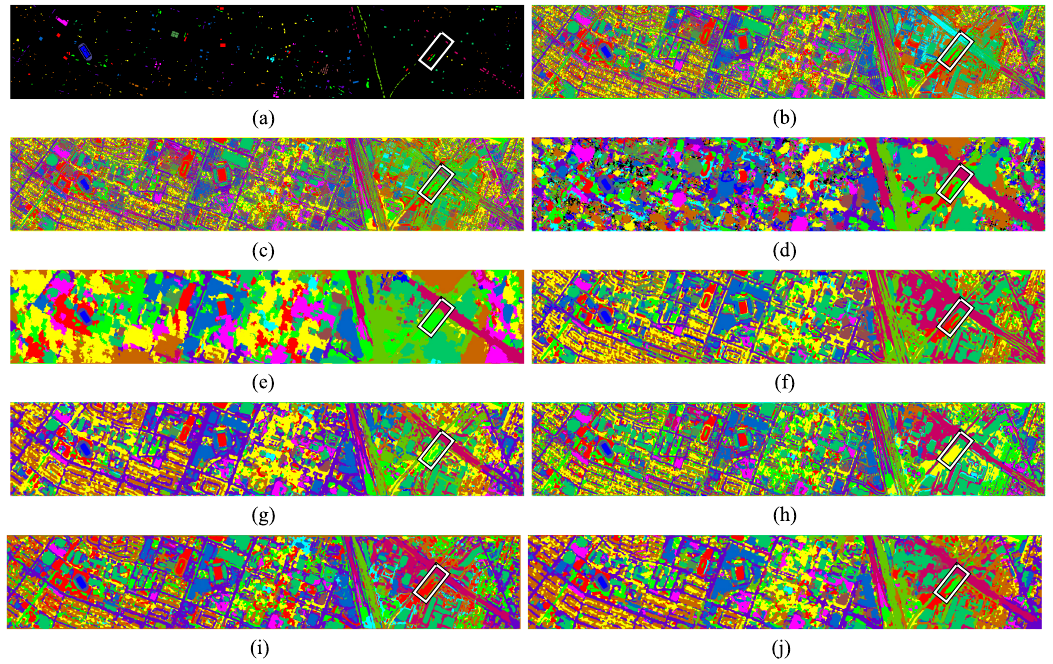}
    \vspace{-5pt}
    \caption{\small{Qualitative visualization of the classification map for Houston dataset. (a) Ground-truth map. (b) SVM. (c) RF. (d) DMSGer. (e) GiGCN. (f) FullyContNet. (g) CLOLN. (h) Spectralformer. (i) GSC-ViT. (j) The proposed MambaHSI. The meaning of colors refers to ~\cref{fig:Houston_dataset} c.}}
    \vspace{-15pt}
    \label{fig:vis_Houston}
\end{figure*}

\begin{figure*}
    \centering
   \includegraphics[width=\linewidth]{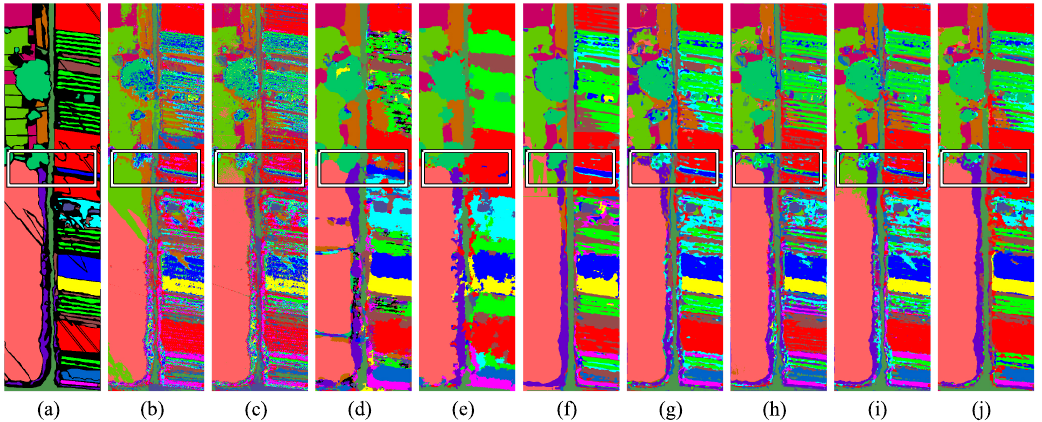}
    \vspace{-5pt}
    \caption{\small{Qualitative visualization of the classification map for HanChuan dataset. (a) Ground-truth map. (b) SVM. (c) RF. (d) DMSGer. (e) GiGCN. (f) FullyContNet. (g) CLOLN. (h) Spectralformer. (i) GSC-ViT. (j) The proposed MambaHSI. The meaning of colors refers to ~\cref{fig:HanChuan_dataset} c.}}
    \vspace{-15pt}
    \label{fig:vis_HanChuan}
\end{figure*}

\begin{figure*}
    \centering
    \includegraphics[width=\linewidth]{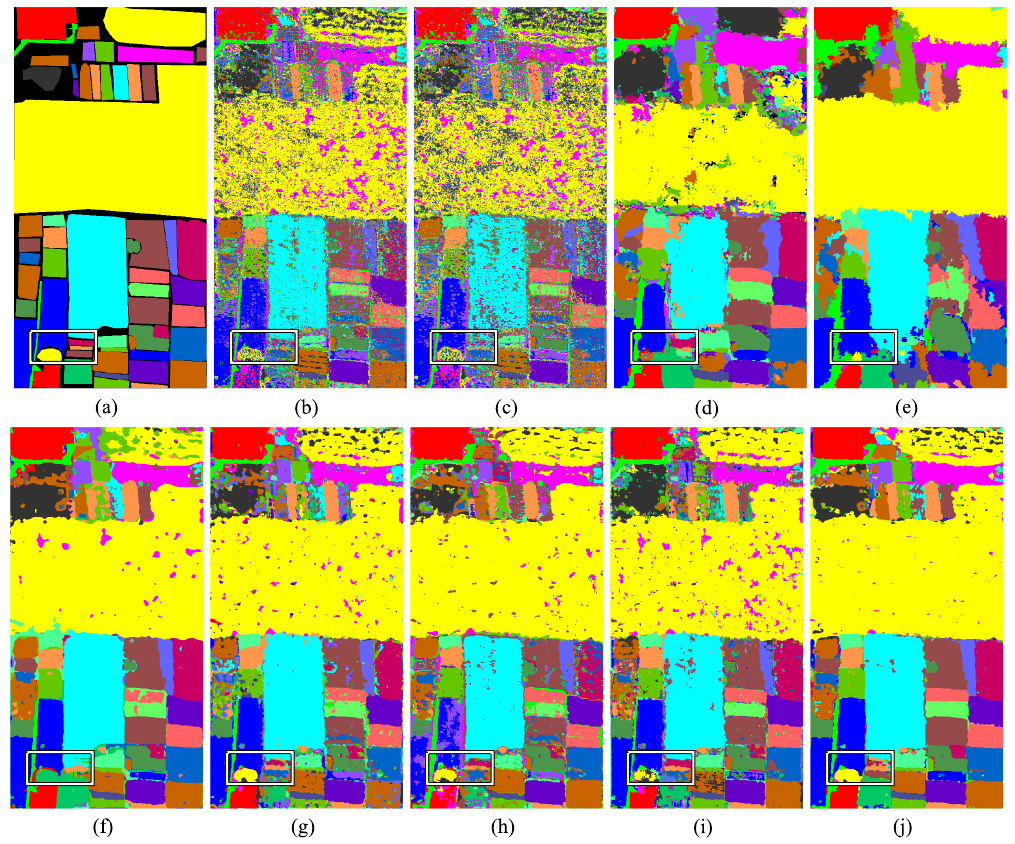}
    \vspace{-15pt}
    \caption{\small{Qualitative visualization of the classification map for HongHu dataset. (a) Ground-truth map. (b) SVM. (c) RF. (d) DMSGer. (e) GiGCN. (f) FullyContNet. (g) CLOLN. (h) Spectralformer. (i) GSC-ViT. (j) The proposed MambaHSI. The meaning of colors refers to ~\cref{fig:HongHu_dataset} c.}}
    \vspace{-15pt}
    \label{fig:vis_HongHu}
\end{figure*}

\subsection{Comparison with State-of-the-Arts}
\blue{We comprehensively compare our method with several state-of-the-arts (SOTA) HSI classification methods from a quantitative and qualitative perspective. These comparison methods cover ML-based~\cite{SVM_Comp,RF_Comp}, GCN-based~\cite{DMSGer,GiGCN}, CNN-based~\cite{FullyContNet,CLOLN} and Transformer-based~\cite{spectralformer,GSC-Vit_TGRS24} methods. The quantitative and qualitative assessments are reported in ~\cref{tab:comp_PaviaU,tab:comp_Houston,tab:comp_HanChuan,tab:comp_HongHu} and  ~\cref{fig:vis_PaviaU,fig:vis_Houston,fig:vis_HanChuan,fig:vis_HongHu}, respectively. From the quantitative results, we find the similar observation with ~\cite{CSIL_ISPRS23}, \textit{i.e.}, the superiority of different backbones is not always prominent, especially in different datasets. Take the Houston dataset as an example, the best transformer-based method GSC-ViT outperforms the best CNN-based CLOLN by $1.54\%$ in overall accuracy (OA). In contrast, the best CNN-based CLOLN gains about $2.32\%$ improvement in OA over the best transformer-based GSC-ViT on Pavia University dataset. Besides, we can observe that the proposed MambaHSI based on SSM outperforms all ML-based, GCN-based, CNN-based and Transformer-based methods, and achieves SOTA on all datasets. This mainly benefits from Mamba's ability to model long-range dependencies with linear computational complexity, which enables MambaHSI to model more detailed long-range dependencies at the pixel level rather than at the patch level. The more detailed pixel-level long-range dependencies simultaneouly provide detailed information (pixel level information) and global conception (long-range dependencies) to boost the discriminability of features for classification. Notably, to prove the potential of the SSM model in HSI classification, we only adopted the basic mamba layer as the basic unit, but did not adopt the newer and stronger mamba visual variant~\cite{VisionMamba} designed for visual images. Even if only the basic mamba layer is used as the basic unit, the performance of the MambaHSI model has exceeded the existing methods based on ML, GCN, CNN and Transformer. This demonstrates the great potential of SSM to be next-generation backbone for hyperspectral image model.}

\blue{From the qualitative results shown in ~\cref{fig:vis_PaviaU,fig:vis_Houston,fig:vis_HanChuan,fig:vis_HongHu}, we can observe that: in comprison with other methods, the proposed MambaHSI has less misclassification, producing a high-quality classification map with smooth objects and well-maintained boundaries. Take the HongHu dataset shown in ~\cref{fig:vis_HongHu} as example, SVM, RF, Spectralformer, GSC-ViT, and CLOLN contain a lot of small area noise. Meanwhile, DMSGer, GiGCN, and FullyContNet have large misclassification areas and fuzzy boundaries. In comparison with these methods, the classification map of the proposed method has better object integrity and clearer boundaries.}

\blue{In summary, when compared with ML-based, GCN-based, CNN-based and Transformer-based methods, the proposed MambaHSI based on SSM can achieve the best classification performance in accuracy and visualization. This reveals the great potential of SSM to be the next-generation backbone for hyperspectral image models.}

\subsection{Ablation Study}
We conduct an ablation study to analyze the effect of each component, including spatial Mamba block, spectral Mamba block, and spatial-spectral fusion module. As shown in ~\cref{tab:ablation_study}, we can observe that: 1) the performance of spatial features exceeds that of spectral features, which indicates that spatial features are more discerning than spectral features; 2) directly summing spatial and spectral information does not always improve model performance, such as the performance degradation in Pavia University and HongHu datasets; 3) after integrating spatial and spectral information by the proposed spatial-spectral fusion module, the model performance is significantly enhanced. Taking HongHu dataset as example, the performance adding SSFM gains $5.47\%$ improvement in OA over spatial information (SpaMB). This demonstrates the effectiveness of the proposed spatial-spectral fusion module, and reveals that spatial features and spectral features are complementary.

\begin{table}[!t]
\setlength{\tabcolsep}{0.5mm}
\begin{tabular}{llll|llll}
\hline
\rowcolor[HTML]{EFEFEF} 
\multicolumn{1}{c}{SpaMB} & \multicolumn{1}{c}{SpeMB} & \multicolumn{1}{c}{Sum} & \multicolumn{1}{c|}{SSFM} & \multicolumn{1}{c}{PaviaU} & \multicolumn{1}{c}{Houston} & \multicolumn{1}{c}{HanChuan} & \multicolumn{1}{c}{HongHu} \\ \hline \hline

\cmark{} & & & & 95.63±0.95 & 92.56±1.31 & 87.84±3.85 & 89.11±1.88  \\
 & \cmark{} & & & 78.34±4.00 & 81.33±2.55 & 73.19±2.79 & 66.97±4.41 \\
\cmark{} & \cmark{} & \cmark{} & & 95.33±1.22 & 92.71±1.58 & 89.29±1.31 & 87.83±2.82 \\
\cmark{} & \cmark{} & & \cmark{} & 95.74±0.90 & 94.46±0.83 & 90.21±1.67 & 94.58±1.01 \\ \hline              
\end{tabular}
\caption{\small{Ablation study of key components of our method. SpaMB, SpeMB, Sum and SSFM denote the spatial Mamba block, spectral Mamba block, summation of spatial and spectral information, and spatial-spectral fusion module, respectively.}}
\vspace{-15pt}
\label{tab:ablation_study}
\end{table}

\blue{\subsection{Complexity Analysis}}
\begin{table*}[]
\setlength{\tabcolsep}{2.3mm}
\begin{tabular}{l|cc|cc|cc|cc|c}
\hline
\rowcolor[HTML]{EFEFEF} 
 & \multicolumn{2}{c|}{ML-based}           & \multicolumn{2}{c|}{GCN-based}          & \multicolumn{2}{c|}{CNN-based}          & \multicolumn{2}{c|}{Transformer-based} & SSM-based                 \\ \cline{2-10} 
\rowcolor[HTML]{EFEFEF} 
 & SVM$^\dag$                 & RF$^\dag$                & DMSGer$^\ddag$               & GiGCN$^\ddag$            & FullyContNet$^\ddag$     & CLOLN$^\dag$                & Spectralformer$^\dag$     & GSC-ViT$^\dag$            & MambaHSI$^\ddag$                  \\ 
 \rowcolor[HTML]{EFEFEF} 
\multirow{-3}{*}{Metrics} & \tiny{TGRS2004} & \tiny{TGRS2005} & \tiny{TNNLS2022} & \tiny{TNNLS2022}                                 & \tiny{TGRS2022} & \tiny{TGRS2024} & \tiny{TGRS2021} & \tiny{TGRS2024}                                 & \tiny{Ours}  \\ \hline\hline 
$\rm{Paramters}$ (M)      & \textbf{-}     & \textbf{-}     & 0.090        & 0.223        & 1.200        & 0.005        & 0.177            & 0.153            & 0.412                        \\
$\rm{FLOPs_{image}}$ (G)   & \textbf{-}     &   \textbf{-}   & 7.66     & 11.01    & 145.62   & 124.44  & 9086.19     & 2026.298    & 39.18                  \\ \hline
\end{tabular}
\vspace{-5pt}
\caption{\small{Comparison of computation cost on the Pavia University dataset. $\rm{FLOPs_{image}}$ denotes the FLOPs to test a whole hyperspectral image. $M$ and $G$ signify mega and giga, respectively. $\dag$ and $\ddag$ denote the patch-level method and image-level approach, respectively.}}
\vspace{-5pt}
\label{tab:Comp_para}
\end{table*}

\begin{table*}[]
\setlength{\tabcolsep}{3mm}
\begin{tabular}{l|cc|cc|cc|cc|c}
\hline
\rowcolor[HTML]{EFEFEF} 
 & \multicolumn{2}{c|}{ML-based}           & \multicolumn{2}{c|}{GCN-based}          & \multicolumn{2}{c|}{CNN-based}          & \multicolumn{2}{c|}{Transformer-based} & SSM-based                 \\ \cline{2-10} 
\rowcolor[HTML]{EFEFEF} 
 & SVM$^\dag$                 & RF$^\dag$                & DMSGer$^\ddag$               & GiGCN$^\ddag$            & FullyContNet$^\ddag$     & CLOLN$^\dag$                & Spectralformer$^\dag$     & GSC-ViT$^\dag$            & MambaHSI$^\ddag$                  \\ 
 \rowcolor[HTML]{EFEFEF} 
\multirow{-3}{*}{Metrics} & \tiny{TGRS2004} & \tiny{TGRS2005} & \tiny{TNNLS2022} & \tiny{TNNLS2022}                                 & \tiny{TGRS2022} & \tiny{TGRS2024} & \tiny{TGRS2021} & \tiny{TGRS2024}                                 & \tiny{Ours}  \\ \hline\hline 
$T_{tr}$ (s)     & 0.01  & 0.29  & 576.88        & 556.31        & 612.15        & 38.43         & 94.19             & 62.21             & 274.54                        \\
$T_{te}$ (s)      & 3.52  & 1.96  & 0.09          & 3.78         & 0.09          & 46.46         & 62.43             & 49.44             & 0.03                          \\ \hline
\end{tabular}
\vspace{-5pt}
\caption{\small{Comparison of running time on the Pavia University dataset. $T_{tr}$ denotes the training time, while $T_{te}$ denotes the time to test a whole hyperspectral image. $s$ denotes seconds. $\dag$ and $\ddag$ denote the patch-level method and image-level approach, respectively.}}
\vspace{-8pt}
\label{tab:Comp_time}
\end{table*}

\begin{table}[!t]
\setlength{\tabcolsep}{0.7mm}
\begin{tabular}{c|cc|cccc}
\hline
\rowcolor[HTML]{EFEFEF} 
Model & Mamba & Self-Attention & $25\times25$   & $50\times50$   & $100\times100$ & $200\times200$ \\ \hline \hline
EB    &   \cmark{}  &               & 0.08  & 0.32 & 1.28   & 5.13   \\
EB    &             &   \cmark{}    & 0.28 & 3.54  & 52.87  & 830.66 \\ \hline
\end{tabular}
\caption{\small{Quantitative comparison of computational complexity between Mamba and Transformer. The GFLOPs are reported.}}
\vspace{-8pt}
\label{tab:mamba_vs_SA}
\end{table}

\begin{figure}[!t]
    \centering
    \includegraphics[width=\linewidth]{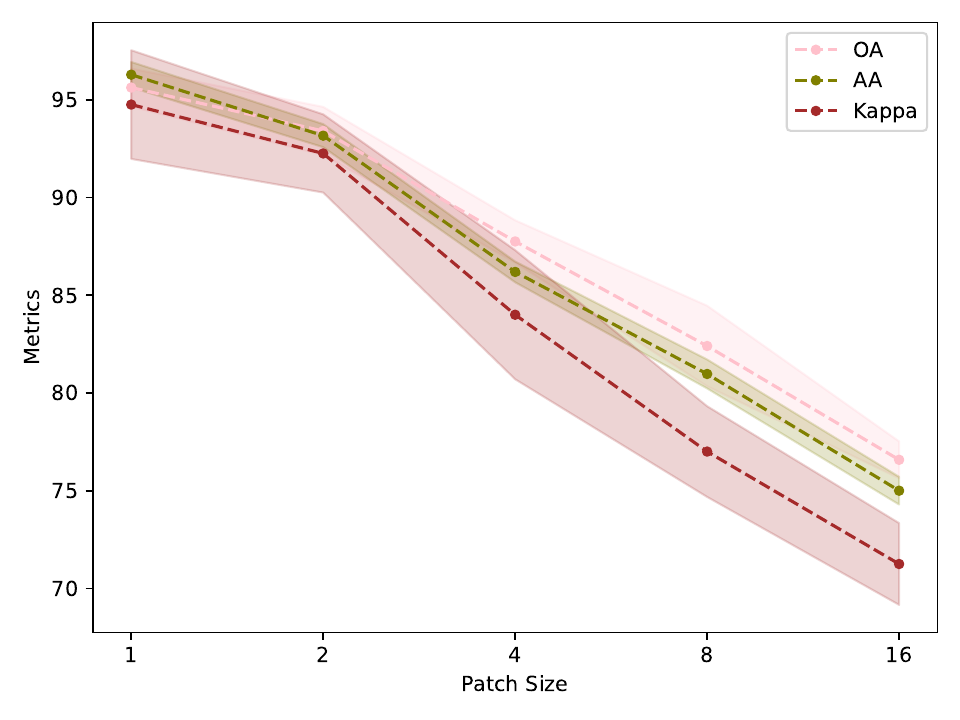}
    \caption{\small{Effect of replacing pixel-wise spatial features with patch features of different sizes on the Pavia University dataset.}}
    \label{fig:effect_of_pixel}
\end{figure}

\blue{We compared the runtime, parameters, and floating point operations (FLOPs) of our method with several SOTA methods on the Pavia University dataset. From the results in~\cref{tab:Comp_para}, we observed that patch-level methods exhibit significantly higher computational complexity for full image processing compared to image-level methods. This is primarily due to the substantial redundant computations occurring between adjacent patches~\cite{SSFCN}. As shown in~\cref{tab:Comp_time}, these redundant computations also result in patch-level methods being significantly slower than image-level methods when testing on full images. Additionally, our method achieved the fastest testing and training times among image-level methods.}
\blue{Furthermore, we quantitatively compared the computational complexity of self-attention and Mamba. Specifically, we generated images of varying sizes and inputted them into an encoder block (EB) composed of a spatial Mamba module (SpaMB), a spectral Mamba module (SpeMB), and a spatial-spectral fusion module (SSFM) to analyze the impact of image sequence length on Mamba's computational complexity. Subsequently, we replaced the Mamba layers in SpaMB and SpeMB with self-attention layers to further analyze the impact of image sequence length on the computational complexity of self-attention layers. As shown in~\cref{tab:mamba_vs_SA}, we can observe that when the model's width and height are each doubled, resulting in a fourfold increase in image sequence length, the computational complexity of the encoder block using the Mamba layer also increases by approximately four times. This quantitatively demonstrates the linear complexity of Mamba. Additionally, when the image size increased from $100\times100$ to $200\times200$, the sequence length increased fourfold. Using an encoder block with self-attention layers increased the computational complexity by approximately sixteenfold. This quantitatively demonstrates the quadratic complexity of self-attention layers.}

\blue{\subsection{Effect of Pixel-wise Spatial Features}}

\blue{To analyze the effect of pixel-wise spatial features, we replaced the pixel-wise features with the patch features of different patch sizes to train MambaHSI and compared them with pixel-wise features. From the results in~\cref{fig:effect_of_pixel}, we can observe that as the patch size increases, the model performance degrades significantly. This demonstrates the importance of pixel-wise features.}

\blue{\subsection{Effect of Spectral Sequence Information}}
\begin{table}[!t]
\setlength{\tabcolsep}{4.8mm}
\begin{tabular}{c|ccc}
\hline
\rowcolor[HTML]{EFEFEF} 
Models & OA & AA & Kappa \\ \hline \hline
w/o SSI    &   75.10±2.31 & 84.54±0.92 & 83.00±3.76   \\
w SSI    &  80.67±2.07 & 88.29±1.21 & 88.50±4.06 \\ \hline
\end{tabular}
\caption{\small{Effect of spectral sequence information (SSI) on the Pavia University dataset.}}
\label{tab:SSI}
\end{table}
\blue{To evaluate the effect of the spectral sequence information, the group $G$ of spectral mamba is set to 1. Because there is only one group ($G=1$), no sequence order relationship between spectra is introduced. As shown in~\cref{tab:SSI} compared to the spectral mamba with introduced sequential spectral information, the Spectral Mamba without spectral sequence information shows performance degradation of $5.57\%$, $3.75\%$, and $5.50\%$ in terms of OA, AA, and Kappa, respectively. This indicates that introducing the spectral sequence information is beneficial for HSI classification.}

\blue{\subsection{Hyper-Parameter Analysis}}
\begin{figure}[!t]
    \noindent \begin{minipage}[]{\linewidth}
    \centering
    \subfloat[]{\includegraphics[width=0.48\linewidth]{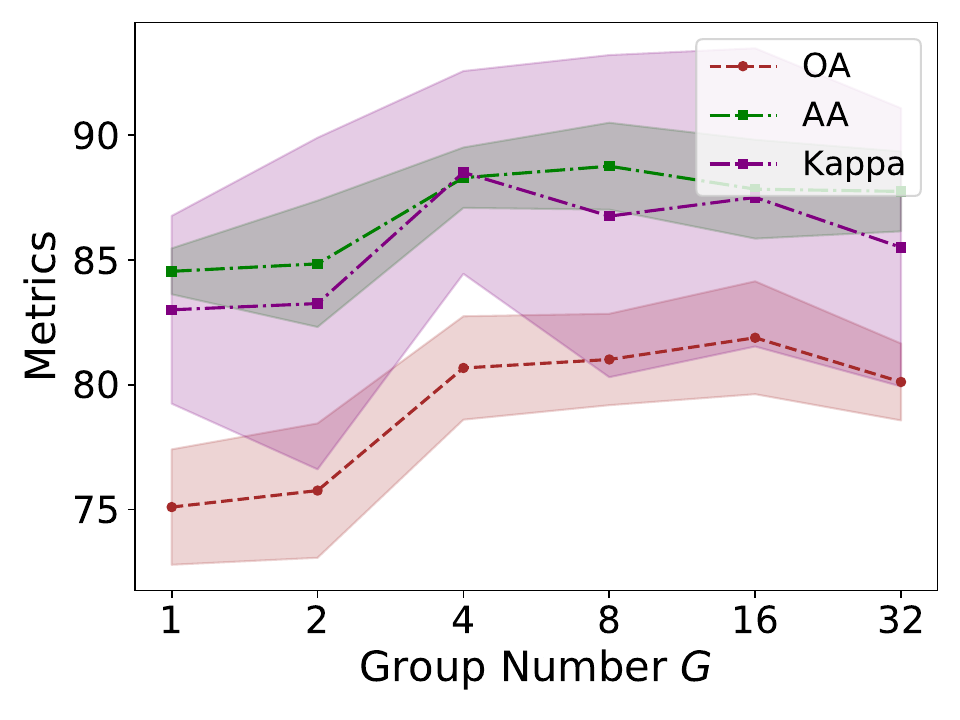}}
    \subfloat[]{\includegraphics[width=0.48\linewidth]{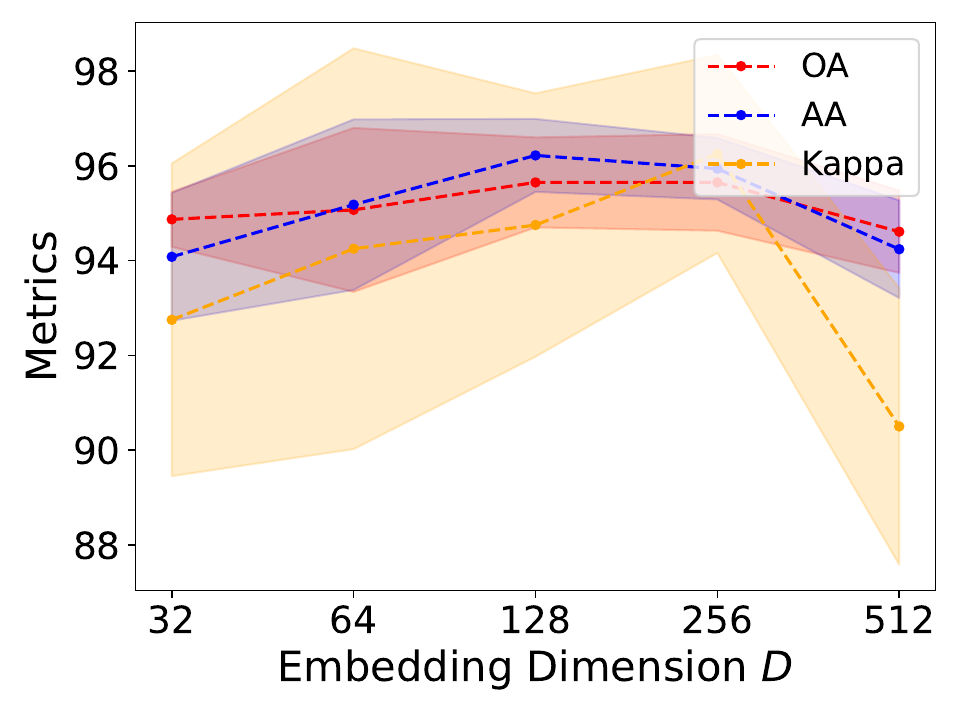}}
    \end{minipage}
    \hfill
    \vspace{-5pt}
    \caption{\small{Hyper-parameter analysis on the Pavia University dataset. (a) the effect of group number G, (b) the effect of embedding dimension D. The figure reports the mean and standard deviation of the results from five runs.}}
    \vspace{-15pt}
    \label{fig:HyperAna}
\end{figure}
\blue{We conducted the hyper-parameter analysis on the Pavia University dataset. From the results in~\cref{fig:HyperAna} (a), we can observe that the performance of Spectral Mamba is the worst when the number of groups for Spectral Mamba is set to 1. This proves the effectiveness of introducing spectral continuity by sequentially inputting multiple groups of spectra into Mamba. Additionally, when the number of spectrum groups is greater than 4, the performance becomes relatively stable. From the results in~\cref{fig:HyperAna}(b), we find that as the number of hidden dimensions increases, model performance initially improves and then declines. Initially, the increase in dimensions enhances the model's learning capacity, resulting in improved performance. However, as the dimensions continue to increase, the sample size of the hyperspectral dataset becomes insufficient to support the model's learning capacity, leading to overfitting and a subsequent decline in performance.}

\section{Conclusion}
In this paper, we present MambaHSI, the first SSM-based image-level HSI classification method, which has strong capability of modeling long-range dependencies and integrating spectral-spatial information. MambaHSI introduces Mamba as a basic unit to model the long-range interactions of the whole image, which enables the model to capture the long-range dependencies of the whole HSI image while maintaining the linear computational complexity. The proposed spatial Mamba block, spectral Mamba block and spatial-spectral fusion module enable the model to mine discriminative spatial and spectral features and then adaptively fuse them for HSI classification.
From the extensive experimental results on multiple datasets, we mainly conclude that: 1) SSM has great potential to be the next-generation backbone for HSI classification benefiting from the strong capability of modeling long-range dependencies while maintaining linear computational complexity; 2) Integrating spatial and spectral information is vital for hyperspectral image classification.
In the future, we intend to extend the proposed idea into more HSI tasks, such as weakly supervised HSI classification and HSI clustering.

{\small
\bibliographystyle{IEEEtran}
\bibliography{MambaHSI}
}

\newpage

\vfill

\end{document}